\documentclass[12pt, a4paper]{article}

\pdfoutput=1
\usepackage{graphicx}
\graphicspath{{./Figures/}}
\usepackage{subfigure}

\usepackage{amssymb,latexsym}

\usepackage{setspace}

\usepackage{amsmath}

\usepackage{mathrsfs,amsfonts}

\usepackage{amsthm,amsxtra}

\usepackage[final]{showkeys}

\usepackage{stmaryrd}

\usepackage{algorithm}
\usepackage{algorithmicx}
\usepackage{algpseudocode}
\usepackage{mathtools}

\usepackage{yhmath}
\usepackage[authoryear,square]{natbib}

\usepackage{amsmath,amssymb,amsthm,mathtools}
\usepackage{microtype,booktabs,tabularx,longtable,multirow}

\usepackage[debug,pdftex,colorlinks=true, 
linkcolor=blue, bookmarksopen=false,
plainpages=false,pdfpagelabels]{hyperref}
\hypersetup{colorlinks=false,
pdfborder={0 0 1},
citebordercolor={0 1 0}}
\usepackage{fancyhdr}

\newtheorem{theorem}{Theorem}[section]
\newtheorem{lemma}[theorem]{Lemma}

\newtheorem{corollary}[theorem]{Corollary}

\DeclareMathOperator*{\argmin}{\arg\min}

 \newcommand{\bbE}{\mathbb E}
 \newcommand{\bbN}{\mathbb N}
\newcommand{\bbP}{\mathbb P} 
\newcommand{\bbR}{\mathbb R}

 \newcommand{\cD}{\mathcal D} 
 \newcommand{\cF}{\mathcal F}
\newcommand{\cG}{\mathcal G} \newcommand{\cH}{\mathcal H}

\newcommand{\cO}{\mathcal O}  
 \newcommand{\cR}{\mathcal R}

 \newcommand{\cZ}{\mathcal Z}

\newcommand{\aver}[1]{\langle {#1} \rangle}

\allowdisplaybreaks

\makeatletter
\newcommand{\chapterauthor}[1]{%
	{\parindent0pt\vspace*{-25pt}%
		\linespread{1.1}\large\scshape#1%
		\par\nobreak\vspace*{35pt}}
	\@afterheading%
}
\makeatother

\title{Improved last-iterate bounds for SGD with and without replacement}
\author{Yimin Zhong~\thanks{Department of Mathematics and Statistics, Auburn University, Auburn, AL 36830;
yzz0025@auburn.edu}\and Jiaxin Zhou~\thanks{Department of Mathematics and Statistics, Auburn University, Auburn, AL 36830;
jiz0004@auburn.edu}}
\date{}

\begin{document}

\maketitle

\begin{abstract}
In this work, we study last-iterate convergence bounds for stochastic gradient descent (SGD) on convex and $\beta$-smooth loss functions and improve the bound from $\cO((\log T)^3/T)$ to $\cO((\log T)^3/(T\log\log T))$ in the interpolation regime. The improvement comes from applying a horizon-dependent multiscale estimate of gradients to the comparison of trajectories. The bound applies to both sampling with replacement and sampling without replacement within one pass.
\end{abstract}


\section{Introduction}
Let $f(\,\cdot\,; z): \bbR^d\to\bbR$ be a family of convex, $\beta$-smooth loss functions, indexed by $z\in [n]$. Consider the optimization problem of the following form
\begin{equation*}
    \min_{x\in\bbR^d} F(x) := \bbE_{z\sim \cZ} \left[ f(x; z) \right],\qquad \cZ = {\rm Unif}([n]).
\end{equation*}
In this note, we study the convergence of the stochastic gradient descent (SGD) algorithm starting from $x_0 \in \bbR^d$:
\begin{equation}
    x_{t} = x_{t-1} - \eta \nabla f(x_{t-1}; z_t), \qquad t=1,\dots, T. 
\end{equation}
The indices $z_1,\dots, z_T$ are either sampled independently from $\cZ$, or chosen as the first $T$ entries of a uniform random permutation. These two strategies are known as \textit{with-replacement} (WR) and \textit{without-replacement} (WOR) samplings. A central question is which sampling rule can yield a smaller expected last-iterate loss with the same problem setting. This comparison motivated extensive studies on the noncommutative AM-GM inequality since~\citet{RechtRe2012}. In this work, we focus on the last-iterate estimates for both sampling strategies. For WOR sampling, we only consider the expected loss within one pass.

Recently, ~\citet{SchlisermanKoren2026} have sharpened the last-iterate bound for WR to nearly optimal $\cO((\log T)^3/T)$ in the interpolation regime. These developments raise the question of what WOR can guarantee 
under the same problem setting. In this work, we use a common framework for WR and WOR with $\eta\in (0, 2\beta^{-1})$, and slightly improve the result to $\cO((\log T)^3/(T\log\log T))$ for the expected loss.

\subsection{Summary of contributions}
Theorem~\ref{thm: main} provides a last-iterate upper bound of order $\frac{(\log T)^3}{T\log\log T}(\frac{R^2}{\delta^3 \eta T} + \frac{\eta\sigma_{\ast}^2}{\delta^4})$ for every stepsize $\eta\in(0, 2\beta^{-1})$ with $\delta = \min\{1, 2-\eta\beta\}$, with noise level $\sigma_{\ast}$ at a global minimizer $x^{\ast}$. The bound applies to both WR and WOR samplings. Our dyadic gradient estimate follows the same idea as~\citet{SchlisermanKoren2026}. However, instead of using the harmonic suffix-averaging technique by~\citet{ShamirZhang} directly, we first apply a comparison between the suffix averages with geometrically increasing separations, which brings an additional saving of $\log\log T$.

\subsection{Related work}
\noindent {\bf Last-iterate SGD bounds.} The last-iterate convergence bound of SGD has been studied extensively. For instance, smooth convex cases (including quadratic loss) are considered in~\citet{LiuZhou, Attia, Garrigos, Evron, Varre, DerezinskiDong, DerezinskiDongCOLT, SchlisermanKoren2026}. 
For average-iterate bounds, see~\citet{Lan, Srebro, CortildLyapunov, BachMoulines}. In particular, ~\citet{DerezinskiDong} establish the optimal last-iterate SGD bound $\cO(1/T)$ in the interpolation regime for convex quadratic loss under WR sampling for stepsizes no larger than the greedy stepsize $\eta=\beta^{-1}$. Their proof can be easily adapted to WOR sampling. ~\cite{SchlisermanKoren2026} generalizes their result to the full range of stepsizes $\eta\in (0, 2\beta^{-1})$ and also achieves a nearly optimal bound $\cO((\log T)^3/T)$ for general cases. Our analysis slightly improves it under the same configuration.
\\\\
\noindent {\bf Random Reshuffling. }
Random Reshuffling (RR) is a common technique in SGD. ~\citet{LiuReshuffling2026, LiuReshufflingExtended2026} analyzed weighted averages under RR. For last-iterates, \citet{ChenZhang2026} give a noisy RR convergence bound under a stepsize of order $1/(\beta\log(nK+2)) $, where $K$ is the number of epochs. ~\cite{LiuGuoHan2026} allow nonincreasing epoch stepsizes under a polylogarithmic constraint. The result in this work considers all iterates within one pass and allows stepsize $\eta\in(0, 2\beta^{-1})$. 
\\\\
\noindent {\bf Noncommutative AM-GM inequality.} ~\citet{RechtRe2012} showed that comparisons of sampling rules are related to the noncommutative AM-GM inequalities. The unrestricted conjecture was disproved by~\citet{LaiLim2020, DeSa2020} with explicit counterexamples. More recently, ~\citet{Peng2026, YunSraJadbabaie2021} studied the case where all matrices are nearly the identity. ~\cite{ZhongAMGM2026} proved the noncommutative AM-GM inequality if $n\ge 2 \lceil m/2\rceil^2$, where $m$ is the product length, which gives a guarantee that WOR can outperform WR under specific conditions.

\section{Main results}
In this section, we present our main results on last-iterate bounds for SGD in the interpolation and low-noise regimes under WR and WOR sampling. 

\subsection{Problem setup}\label{sec: pro}
Let $\|\cdot\|_2$ be the Euclidean vector norm.
For every $z\in Z =[n]$, we assume $f(\,\cdot\,; z):\bbR^d\to \bbR$ is convex, and for some $\beta > 0$, 
\begin{equation*}
    \|\nabla f(x; z) - \nabla f(y; z)\|_2 \le \beta \|x - y\|_2,\qquad \forall x, y\in\bbR^d.
\end{equation*}
The problem to be considered is an unconstrained optimization: 
\begin{equation*}
    \min_{x\in\bbR^d} F(x),\qquad F(x) := \bbE_{z\sim\cZ} f(x; z).
\end{equation*}
Assume $F$ permits a global minimizer, let $x^{\ast}\in\argmin_{x\in\bbR^d} F(x)$, and define 
\begin{equation}
    R := \|x_0 - x^{\ast}\|_2 ,\qquad \sigma_{\ast}^2 := \bbE_{z\sim \cZ} \|\nabla f(x^{\ast}; z)\|_2^2,\qquad \cG(x) = \bbE_{z\sim \cZ} \|\nabla f(x; z)\|_2^2.
\end{equation}
where $\sigma_{\ast}^2$ quantifies the gradient variance at $x^{\ast}$, we assume $\sigma_{\ast}$ is bounded. The case $\sigma_{\ast} = 0$ is called the \textit{interpolation regime}, where $x^{\ast}$ is the shared minimizer for all $f(\,\cdot\,;z)$ across all $z\in[n]$. The case $\sigma_{\ast} \neq 0$ is called the \textit{low-noise regime}.

Starting from a deterministic $x_0\in\bbR^d$, the SGD follows
\begin{equation*}
    x_{t} = x_{t-1} - \eta \nabla f(x_{t-1}; z_t),\qquad t=1,\dots, T,
\end{equation*}
where $\eta > 0$ is the learning rate and $T$ counts the number of updates. The last iterate is $x_{T}$. 

In this note, two common sampling strategies for $\{z_t\}_{t=1}^T$ are considered: 
Under WR sampling, $z_1, \dots, z_T$ are independently drawn from $\cZ$, and $T\ge 1$ is arbitrary; Under WOR sampling, $z_1,\dots, z_T$ are the first $T$ entries of a random permutation uniformly drawn from the symmetric group $\mathfrak{S}(n)$ with $1\le T\le n$.

\subsection{Main results}
For $s\ge 1$, we define 
\begin{equation}\label{eq: log}
    \ell_s := \log(s+2), \qquad \lambda_s := \log(\ell_s + 2).
\end{equation}
In particular, we note that $\lambda_T = \Omega(\log\log T)$ for large $T$. 
We state our main results below. 
\begin{theorem}\label{thm: main}
Fix a constant $\eta\in(0, 2\beta^{-1})$, and let 
\begin{equation*}
    \delta = \min\{1, 2 - \eta \beta\},\qquad P_{s} = \frac{\ell_{s}^3}{\lambda_{s}}. 
\end{equation*}
Then under WR for every $T\ge 1$, and under WOR for every $1\le T\le n$, 
\begin{equation*}
    \bbE[F(x_{T}) - F(x^{\ast})] \le C P_T \left[  \frac{R^2}{\delta^3\eta T} + \frac{\eta \sigma_{\ast}^2 }{\delta^4}  \right],
\end{equation*}
where $C$ is an absolute constant. 
\end{theorem}
\begin{corollary}\label{thm: inter}
    Assume that $\sigma_{\ast} = 0$ and use $\eta = \beta^{-1}$. Then for WR sampling at every $T \ge 1$, and for WOR sampling at every $1 \le T \le n$, 
\begin{equation*}
    \bbE[F(x_T) - F(x^{\ast})] \le \frac{C \beta R^2 \ell_T^3}{T \lambda_T},\qquad    \bbE[\cG(x_{T})] \le \frac{C\beta^2  R^2 \ell_T^2}{T},
\end{equation*}
where $C$ is an absolute constant.
\end{corollary}
Since $\ell_T =\cO(\log T)$, the above last-iterate upper bound with greedy step $\eta=\beta^{-1}$ in the interpolation regime becomes $\cO((\log T)^3/(T\log\log T))$, regardless of WR or WOR sampling, which is sharp up to a polylogarithmic factor and improves the previous upper bound $\cO((\log T)^3/T)$ in~\citet{SchlisermanKoren2026} slightly. In particular, for convex quadratic loss functions, the last-iterate sharp bound $\cO(1/T)$ was recently proved in~\citet{DerezinskiDong} for WR sampling, and their proof can be easily adapted to WOR sampling. However, in general cases, it remains challenging to improve the polylogarithmic factor further. 

The remaining sections are organized as follows. 
In Section~\ref{sec: inter}, we introduce the preliminaries throughout this work and prove Corollary~\ref{thm: inter} in the interpolation regime $\sigma_{\ast} = 0$. The low-noise regime is considered in Section~\ref{sec: noise}. The conclusion is given in Section~\ref{sec: conc}.

\section{Interpolation regime}\label{sec: inter}
We assume $\sigma_{\ast} = 0$ first to explain the core idea. The low-noise regime $\sigma_{\ast} \neq 0$ follows the same strategy. The minimizer $x^{\ast}$ is shared by all $f(x, z)$ for $z\in [n]$. 
Without loss of generality, we assume $f(\,\cdot\,;z)$ is nonnegative, convex, and 1-smooth for every $z\in[n]$. Otherwise, we can replace $f$ with its normalization:
\begin{equation*}
    f(x, z) \longleftarrow \frac{1}{\beta} (f(x, z) - f(x^{\ast}, z)). 
\end{equation*}
In the normalized case, the stepsize $\eta = \frac{1}{\beta} = 1$, and the SGD follows
\begin{equation*}
    x_{t} = x_{t-1} - \nabla f(x_{t-1}; z_t),\qquad t=1,\dots, T.
\end{equation*}

\subsection{Preliminaries}
We first introduce a few preliminary estimates. The following lemma is standard and shows that the SGD update is non-expansive. This is an important property for later use. 
\begin{lemma}\label{lem: nonex}
Let $h$ be a convex, 1-smooth function. Then for any $u, v\in\bbR^d$,   
\begin{equation}\label{eq: T}
    \|(u - \nabla h(u)) - (v - \nabla h(v))\|_2^2 +\|\nabla h(u) - \nabla h(v)\|_2^2 \le \|u - v\|_2^2.
\end{equation}
Moreover, if $h\ge 0$, then $\|\nabla h(u)\|_2^2 \le 2 h(u)$ for any $u\in\bbR^d$.
\end{lemma}
\begin{proof}
By the 1-smoothness and co-coercivity, 
\begin{equation}\label{eq: conv}
\aver{\nabla h(u) - \nabla h(v), u - v} \ge  \|\nabla h(u) -\nabla h(v)\|_2^2.
\end{equation}
Rearranging~\eqref{eq: conv} gives
\begin{equation}
    \|(u - \nabla h(u)) - (v - \nabla h(v))\|_2^2 + \|\nabla h(u) - \nabla h(v)\|_2^2 \le \|u - v\|_2^2.
\end{equation}
In particular, if $h$ is nonnegative, 
\begin{equation}\label{eq: low}
\begin{aligned}
    0\le h(u - \nabla h(u)) &\le h(u) - \aver{\nabla h(u), \nabla h(u)} + \frac{1}{2} \|\nabla h(u)\|_2^2 \\&=  h(u) - 
    \frac{1}{2}\|\nabla h(u)\|_2^2,
\end{aligned}
\end{equation}
which gives $\|\nabla h(u)\|_2^2 \le 2 h(u)$.
\end{proof}

\begin{lemma}\label{lem: init}
Let $h_1,\dots, h_s$ be a deterministic sequence of nonnegative, convex, 1-smooth functions. Assume the iteration $u_{j} = u_{j-1} - \nabla h_{j}(u_{j-1})$, then
\begin{equation*}
    \|u_{s} - u_0\|_2^2 + \sum_{j=1}^s h_j(u_{j-1}) \le 4 \sum_{j=1}^s h_j(u_0).
\end{equation*}
\end{lemma}

\begin{proof}
    Let $p_{j-1} = u_{j-1}- u_0$ and $q_{j} = \nabla h_j(u_{j-1})$, and $r_j = \nabla h_{j}(u_0)$, then $u_{j} - u_0 = p_{j-1} - q_j$. Using 1-smoothness and convexity,  
    \begin{equation}\label{eq: init loss}
        h_j(u_0) \ge h_j(u_{j-1}) - \aver{q_{j}, p_{j-1}} + \frac{1}{2}\|q_j - r_j\|_2^2. 
    \end{equation}
    Then by~\eqref{eq: init loss},
    \begin{equation*}
    \begin{aligned}
        \|p_{j-1} - q_j\|_2^2 - \|p_{j-1}\|_2^2 &= \|q_j\|_2^2 - 2\aver{p_{j-1}, q_j} \\&\le \|q_j\|_2^2 + 2 \left(   h_j(u_0) - h_j(u_{j-1}) - \frac{1}{2}\|q_j - r_j\|_2^2 \right) \\
        &= 2( h_j(u_0) - h_j(u_{j-1})) + 2 \aver{q_j, r_j} - \|r_j\|^2_2\\
        &= 2( h_j(u_0) - h_j(u_{j-1})) + \|r_j\|_2^2 + \frac{1}{2}\|q_j\|_2^2  - \frac{1}{2} \|q_j - 2 r_j\|_2^2 \\
        &\le 2 h_j(u_0) - h_j(u_{j-1}) + 2 h_j(u_0) - \frac{1}{2} \|q_j - 2 r_j\|_2^2\\
        &\le  4 h_j(u_0) - h_j(u_{j-1}), 
    \end{aligned}
    \end{equation*}
    where we have applied~\eqref{eq: low} that $\|q_j\|_2^2 \le 2h_j(u_{j-1})$ and $\|r_j\|_2^2 \le 2h_j(u_0)$. Rearranging the above inequality and summing telescopically yields the desired inequality.
\end{proof}
\begin{corollary}\label{cor: init dist}
    Suppose in addition to the assumptions of Lemma~\ref{lem: init}, there exists $u^{\ast}\in\bbR^d$ such that $h_j(u^{\ast}) = 0$ for every $j=1,\dots, s$. Then 
    \begin{equation}\label{eq: bounds}
        \sum_{j=1}^s h_j(u_{j-1}) \le \frac{1}{2}\|u_0 - u^{\ast}\|_2^2,\qquad  \sum_{j=1}^s \|u_{j-1} - u_{j}\|_2^2 \le \|u_0 - u^{\ast}\|_2^2. 
    \end{equation}
\end{corollary}
\begin{proof}
    Since $h_j$ is nonnegative, $u^{\ast}$ is a global minimizer of $h_j$, thus $\nabla h_j(u^{\ast}) = 0$. Set
    \begin{equation*}
        \theta_{j-1} = u_{j-1} - u^{\ast},\qquad q_j = \nabla h_j(u_{j-1}). 
    \end{equation*}
    Using 1-smoothness and convexity, we obtain
    \begin{equation}\label{eq: hj neg}
        0 = h_j(u^{\ast}) \ge h_j(u_{j-1}) - \aver{\theta_{j-1},  q_j}  + \frac{1}{2} \|q_j\|_2^2.
    \end{equation}
    Similar to the derivation in Lemma~\ref{lem: init}, 
    \begin{equation*}
        \|u_{j-1} - u^{\ast}\|_2^2 - \|u_{j} - u^{\ast}\|_2^2 = \|\theta_{j-1}\|^2 - \|\theta_{j-1} - q_j \|_2^2 = 2 \aver{ \theta_{j-1}, q_j} - \|q_j\|_2^2. 
    \end{equation*}
    Combining~\eqref{eq: hj neg}, 
    \begin{equation*}
      2 h_j(u_{j-1})\le  \|u_{j-1} - u^{\ast}\|_2^2 - \|u_{j} -u^{\ast}\|_2^2 .
    \end{equation*}
    Perform a telescoping sum over $j$, then we obtain the first estimate in~\eqref{eq: bounds}:
    \begin{equation}\label{eq: h sum}
        2\sum_{j=1}^s h_j(u_{j-1}) \le \|u_0 - u^{\ast}\|_2^2 - \|u_{s} - u^{\ast}\|_2^2 \le \|u_0 - u^{\ast}\|_2^2.
    \end{equation}
For the other estimate, we use~\eqref{eq: low},
\begin{equation*}
    \|u_{j-1} - u_{j}\|_2^2 = \|\nabla h_j(u_{j-1})\|_2^2 \le 2 h_j(u_{j-1}).
\end{equation*}
A summation over $j$ combined with~\eqref{eq: h sum} gives
\begin{equation*}
    \sum_{j=1}^s \|u_{j-1} - u_{j}\|_2^2 \le 2\sum_{j=1}^s h_j(u_{j-1}) \le \|u_0 - u^{\ast}\|_2^2.
\end{equation*}
\end{proof}

\subsection{Invariant sample space}\label{sec: inv}
Fix a batch length $m\ge 1$. Let $\Omega\subset [n]^m$ be a nonempty set of tuples closed under the cyclic shift
\begin{equation*}
    \cR :\Omega\longrightarrow\Omega,\qquad \cR (z_1,\cdots, z_m) = (z_m, z_1,\dots, z_{m-1}).
\end{equation*}
Equip $\Omega$ with a probability law $\bbP$. We call $(\Omega, \bbP)$ \emph{$\cR$-invariant} if 
\begin{equation*}
    \bbP(\cR^{-1}A) = \bbP(A),\qquad \forall A\subset \Omega.
\end{equation*}
In particular, for uniform sampling with replacement, we take $\Omega = [n]^m$, and for uniform sampling without replacement, we take $\Omega$ to be the set of ordered distinct $m$-tuples with $m\le n$. Both cases are $\cR$-invariant.

Suppose $(\Omega, \bbP)$ is $\cR$-invariant, and we define
\begin{equation*}
    \cH = L^2(\Omega, \bbP; \bbR^d), \qquad \|Y\|_{\cH}^2 := \bbE_{\omega}[\|Y(\omega)\|_2^2].
\end{equation*}
The invariance implies that 
\begin{equation*}
    \|Y\circ \cR\|_{\cH}^2 = \bbE_{\omega}[\|Y(\cR\omega)\|_2^2] = \bbE_{\omega}[\|Y(\omega)\|_{2}^2] = \|Y\|_{\cH}^2.
\end{equation*}
Let $\omega = (z_1, \dots, z_m)\in [n]^m$ be a given tuple. For each $1\le k\le m$, we consider the iteration 
\begin{equation}\label{eq: u update}
    y_0^{(k)} = x_0,\qquad y_{j}^{(k)} = y_{j-1}^{(k)} - \nabla f(y_{j-1}^{(k)}; z_{m-k+j}),\qquad j=1,\dots, k.
\end{equation}
The trajectory $\{y_j^{(k)}\}_{j=1}^k$ consists of the SGD gradient updates through indices $z_{m-k+1},\dots,z_{m}$ with the same deterministic starting point $x_0$. Define 
\begin{equation*}
    X_k(\omega) = y_{k}^{(k)} - x_0\qquad (1\le k\le m),\qquad G_k(\omega) = \nabla f( y_{k}^{(k+1)}; z_m)\qquad (0 \le k < m).
\end{equation*}
Then $X_k(\omega)$ is the final displacement after applying the last $k$ gradient updates from $\omega$. Notice that the last $k$ indices of $\cR\omega$ are $z_{m-k},\dots, z_{m-1}$, thus, $x_0 + X_{k}(\cR\omega) = y_{k}^{(k+1)}$. Starting with $X_0(\omega) = 0$, for $0\le k < m$,  we have
\begin{equation}\label{eq: traj}
\begin{aligned}
     G_{k}(\omega) &= \nabla f(x_0 + X_{k}(\cR\omega); z_m),\\
    X_{k+1}(\omega) &= X_{k}(\cR\omega) - G_k(\omega).
\end{aligned}
\end{equation}
The following lemma bridges~\eqref{eq: traj} to SGD more explicitly.  
\begin{lemma}\label{lem: bridge}
Let $\omega = (z_1,\dots, z_m)\in [n]^m$. Consider the SGD iterations guided by $\omega$:
\begin{equation*}
    x_0(\omega) = x_0,\qquad x_{j} = x_{j-1} - \nabla f(x_{j-1};z_{j}), \qquad 1\le j\le m.
\end{equation*}
Then for every $0\le k\le m$, the trajectory defined by~\eqref{eq: traj} satisfies
\begin{equation}\label{eq: bridge}
\begin{aligned}
    x_0 + X_k(\omega) &= x_{k}(\cR^k\omega).
\end{aligned}
\end{equation}
\end{lemma}
\begin{proof}
Notice that for every $1\le k\le m$, 
\begin{equation*}
    \cR^k \omega = (z_{m-k+1},\dots, z_m, z_1, \dots, z_{m-k}),\qquad (\cR^k\omega)_k  = z_m.
\end{equation*}
We prove~\eqref{eq: bridge} by induction. At $k=0$, the equality holds since $X_0(\omega) = 0$. Assume the equality holds for $k-1$. Then 
    \begin{equation}\label{eq: 2nd bridge}
        x_0 + X_{k-1}(\cR \omega) = x_{k-1}(\cR^{k-1} \cR  \omega) = x_{k-1}(\cR^k \omega),
    \end{equation}
 By~\eqref{eq: traj} and~\eqref{eq: 2nd bridge},
    \begin{equation*}
    \begin{aligned}
        x_0 + X_{k}(\omega) &= x_0 + X_{k-1}(\cR \omega) - \nabla f(x_0 + X_{k-1}(\cR\omega);{z_m})\\
        &=x_{k-1}(\cR^k \omega) -  \nabla f(x_{k-1}(\cR^k \omega) ; {(\cR^k\omega)_k})\\
        &= x_{k}(\cR^k \omega).
    \end{aligned}
    \end{equation*}
    Thus, the induction is complete.
\end{proof}
\begin{corollary}\label{cor: h norm}
    Suppose the sampling law is $\cR$-invariant. Then the trajectories in Lemma~\ref{lem: bridge} satisfy  
    \begin{equation*}
    \begin{aligned}
            \|X_k\|_{\cH}^2 &= \bbE_{\omega}[\|x_{k}(\omega) - x_0\|_2^2], \qquad &&0\le k\le m, \\
            \|G_k\|_{\cH}^2 &= \bbE_{\omega}[\|\nabla f (x_{k}(\omega); z_{k+1} )\|_2^2],\qquad &&0\le k < m.
    \end{aligned}
    \end{equation*}
    and for $0\le k < m$, $$\bbE_{\omega}[f(x_0 + X_{k}(\cR\omega);{z_m})] = \bbE_{\omega}[f(x_{k}(\omega); z_{k+1})].$$
\end{corollary}
\begin{proof}
    Using~\eqref{eq: bridge} and the invariance of law under $\cR$, 
    \begin{equation*}
      \|X_k\|_{\cH}^2 = \bbE_{\omega}[ \|x_{k}(\cR^k\omega) - x_0\|_2^2] =   \bbE_{\omega}[ \|x_{k}(\omega) - x_0\|_2^2]. 
    \end{equation*}
    Similarly, 
    \begin{equation*}
    \begin{aligned}
           \|G_k\|_{\cH}^2 = \bbE_{\omega}[\|\nabla f (x_0 + X_{k}(\cR \omega);{z_m}) \|_2^2 ] &= \bbE_{\omega}[\|\nabla f(x_{k}(\cR^{k+1} \omega);(\cR^{k+1}\omega)_{k+1})\|_2^2]\\
           &= \bbE_{\omega}[\|\nabla f (x_{k}(\omega); z_{k+1})\|_2^2].
    \end{aligned}
    \end{equation*}
    The last one follows the same derivation.
\end{proof}
Next, we estimate an upper bound of $\|G_k\|_{\cH}^2$ for later use. The following Lemma~\ref{lem: nonneg} follows directly from Lemma~\ref{lem: init}, which gives a bound of $\sum_{k=0}^{m-1} \|G_k\|_{\cH}^2$ at the scale of $\cO(m)$.
Then, in Lemma~\ref{lem: diff}, we show that $\sum_{k=s}^{m-1}\|G_k - G_{k-s}\|_{\cH}^2$ varies at most linearly in $s$, which poses a strong constraint on each $\|G_k\|_{\cH}^2$, and prevents it to grow too fast. Importantly, we emphasize that $\|G_k - G_{k-s}\|_{\cH}^2$ is \textit{not} an expectation of $\|\nabla f(x_k, z_{k+1}) - \nabla f(x_{k-s}, z_{k-s+1})\|_2^2$. 
\begin{lemma}\label{lem: nonneg}
    Suppose each component of $f$ is non-negative, convex, and 1-smooth. Let the sampling law be $\cR$-invariant. Then for every $0\le s\le m$,
    $$\|X_s\|_{\cH}^2 \le 4 s A_0,\qquad \sum_{k=0}^{m-1} \|G_k\|_{\cH}^2 \le 8mA_0, $$
    where $A_0 := \bbE_{\omega} f(x_0; z_m) = \bbE_{z\sim\cZ} f(x_0, z)$. 
\end{lemma}
\begin{proof}
    By Lemma~\ref{lem: init}, 
    \begin{equation}\label{eq: X bd}
         \|x_{s}(\omega) - x_0\|_2^2 + \sum_{j=1}^s f(x_{j-1}, z_j) \le 4 \sum_{j=1}^s f(x_0, z_j).
    \end{equation}
    Take the expectation over $\omega\in\Omega$ and use the non-negativity of $f$, we obtain
    \begin{equation*}
        \|X_s\|_{\cH}^2 \le 4 s \bbE_{\omega}[f(x_0, z_m)] = 4s A_0.
    \end{equation*}
    Replace $s$ by $m$ in~\eqref{eq: X bd} and apply~\eqref{eq: low}, then take expectation over $\omega$, we have 
    \begin{equation*}
       \sum_{k=0}^{m-1} \|G_k\|_{\cH}^2 \le 2\sum_{k=0}^{m-1} \bbE_{\omega}[f(x_{k}(\omega); z_{k+1})] \le 8\sum_{k=0}^{m-1} \bbE_{\omega}[f(x_0; z_{k+1})] = 8m A_0.
    \end{equation*}
  
\end{proof}

\begin{lemma}\label{lem: diff}
Under the assumptions of Lemma~\ref{lem: nonneg}, for every $1\le s < m$, 
    \begin{equation}\label{eq: g diff}
        \sum_{k=s}^{m-1} \|G_k - G_{k-s}\|_{\cH}^2 \le \|X_s\|_{\cH}^2 - \|X_{m} - X_{m-s}\|_{\cH}^2 \le 4 s A_0,
    \end{equation}
    where  $A_0 := \bbE_{\omega} f(x_0; z_m) = \bbE_{z\sim\cZ} f(x_0, z)$. 
\end{lemma}
\begin{proof}
Apply~\eqref{eq: T} to $h = f(\,\cdot\,; z_m)$ and
\begin{equation*}
    u = x_0 + X_{k}(\cR\omega), \qquad v = x_0 + X_{k - s}(\cR\omega).
\end{equation*}
Using~\eqref{eq: traj}, we obtain
\begin{equation*}
    u -\nabla h(u) = x_0 + X_{k+1}(\omega), \qquad v - \nabla h(v) = x_0 + X_{k-s+1}(\omega).  
\end{equation*}
Thus, taking the expectation over $\omega\in\Omega$, 
\begin{equation*}
\begin{aligned}
 \|X_{k+1} - X_{k-s+1}\|_{\cH}^2 + \| G_{k} - G_{k-s} \|_{\cH}^2 &\le \bbE_{\omega}[\| X_{k}(\cR\omega) - X_{k - s}(\cR\omega)\|_2^2]\\&= \|X_{k} -X_{k - s} \|_{\cH}^2.
\end{aligned}
\end{equation*}
Summing over $k=s,\dots, m-1$, we obtain
\begin{equation*}
\sum_{k=s}^{m-1} \|G_k - G_{k-s}\|_{\cH}^2 \le \|X_s - X_0\|_{\cH}^2 - \|X_{m} - X_{m-s}\|_{\cH}^2.
\end{equation*}
Applying $X_0 = 0$ and Lemma~\ref{lem: nonneg} proves the ~\eqref{eq: g diff}.
\end{proof}
The following lemma proves an upper bound for $\|G_{m-1}\|_{\cH}^2$ using a dyadic weight. The bound is deliberately non-optimized to reveal the $\cO( (\log m)^2 )$ dependence. A stronger estimate is carried out later in Lemma~\ref{lem: am bound}.
\begin{lemma} \label{thm: grad}
Suppose $f(\,\cdot\,;z)$ is nonnegative, convex, and 1-smooth for every $z\in[n]$, and the sampling law is $\cR$-invariant. Then
    $$\|G_{m-1}\|_{\cH}^2 = \bbE_{\omega}[\|\nabla f(x_{m-1}, z_m)\|_2^2] \le (\lfloor \log_2 m \rfloor + 4)^2 A_0. $$
\end{lemma}
\begin{proof}
    For $1\le l\le m$, we define the average of the last $l$ gradients by
    \begin{equation*}
        U_l(\omega) = \frac{1}{l}\sum_{k = m-l}^{m-1} G_k(\omega).
    \end{equation*}
    In particular, $U_1=G_{m-1}$. Then if $2 l\le m$, 
    \begin{equation*}
        U_l - U_{2l} = \frac{1}{2l} \sum_{k=m-l}^{m-1} (G_k - G_{k-l}).
    \end{equation*}
    Using the Cauchy-Schwarz inequality and combining Lemma~\ref{lem: diff},
    \begin{equation*}
        \| U_l - U_{2l} \|_{\cH}^2 \le \frac{1}{4l} \sum_{k=m-l}^{m-1} \|  G_k - G_{k-l}  \|_{\cH}^2 \le A_0.
    \end{equation*}
    Choose $J = \lfloor \log_2 m \rfloor$, then $m \le 2^{J+1}$. By the Cauchy-Schwarz inequality, 
    \begin{equation*}
       \| U_{2^J} \|_{\cH}^2 \le \frac{1}{2^J} \sum_{k=m - 2^J}^{m-1} \|G_k\|_{\cH}^2 \le \frac{8 m A_0}{2^J} \le 16 A_0.
    \end{equation*}
    Therefore, 
    \begin{equation*}
      \|U_1\|_{\cH} = \|G_{m-1}\|_{\cH} \le \|U_{2^J}\|_{\cH} + \sum_{j=0}^{J-1} \|U_{2^{j}} - U_{2^{j+1}}\|_{\cH} \le (4 + J)\sqrt{A_0}.
    \end{equation*}
\end{proof}

\subsection{Trajectory comparison}\label{sec: tr comp}
In this section, we restrict the sampling law to WR or WOR. After conditioning on the first $s$ indices $z_1,\dots, z_s$, the later indices are either uniformly distributed on $[n]$ or on the remaining $n-s$ indices. We denote $\cF_s = \sigma(z_1,\dots, z_s)$ the $\sigma$-algebra on the first $s$ indices, and let 
\begin{equation*}
    A_s = \bbE_{\omega} f(x_s; z_{s+1}),\qquad S_m = \sum_{s=0}^{m-1} A_s.
\end{equation*}
Lemma~\ref{lem: init} implies $S_m \le 4 m A_0$. The following lemma compares two trajectories with common updates in the tail.
\begin{lemma}\label{lem: com run}
Suppose $f(\cdot; z)$ is non-negative, convex, and 1-smooth for every $z\in [n]$. Fix a length $m$ under WR or WOR. Then for $0 \le r \le s$, $l \ge 1$, and $s + l \le m$. 
\begin{equation}\label{eq: traj comp}
\begin{aligned}
    \sum_{j=s}^{s + l - 1} \left( A_j - \frac{1}{2} \|G_j\|_{\cH}^2 \right) &\le l A
    _r + \frac{1}{2}\|X_{s} - X_r\|_{\cH}^2 \\& \le l A
    _r + 2(s-r) A_0. 
\end{aligned}
\end{equation}
\end{lemma}

\begin{proof}
    Let $\omega = (z_1,\dots, z_m)\in [n]^m$. Recall that in Section~\ref{sec: inv}, $X_k(\omega)$ is the final displacement after applying the last $k$ gradient updates from $\omega$. Therefore, the last $r$ updates of $X_s(\omega)$ and $X_r(\omega)$ are the shared $(z_{m-r+1}, \cdots ,z_m)$. Before applying these shared updates, the two displacements are $X_{s-r}(\cR^r\omega)$ and $X_0 = 0$. Because of the non-expansiveness (Lemma~\ref{lem: nonex}), 
    \begin{equation*}
        \|X_{s}(\omega) - X_r(\omega)\|^2_2 \le \|X_{s-r}(\cR^r\omega)\|_2^2.
    \end{equation*}
    By Lemma~\ref{lem: nonneg}, after taking the expectation over $\omega$ and use the $\cR$-invariance, 
    \begin{equation}\label{eq: s-r}
        \|X_s - X_r\|_{\cH}^2 \le \|X_{s-r}\|_{\cH}^2 \le 4 (s-r) A_0.
    \end{equation}
    Define the trajectory for comparison,
    \begin{equation*}
        Y_j(\omega) = X_r(\cR^{j-s}\omega),\qquad s\le j \le s+l.
    \end{equation*}
    Then $Y_{j+1}(\omega) = Y_j(\cR \omega)$. For $s \le j < s + l$, let $u = j - s + 1$, then $u\ge 1$ and $r + u\le m$. We apply Lemma~\ref{lem: bridge},
    \begin{equation*}
        x_0 + X_r(\cR^u \omega) = x_r(\cR^{r+u}\omega),\qquad (\cR^{r+u}\omega)_{r+u} = z_m.
    \end{equation*}
    Therefore, using the $\cR$-invariance, 
    \begin{equation*}
        \bbE_{\omega} f(x_0 + Y_j(\cR \omega); z_m) =  \bbE_{\omega} f(x_r(\cR^{r+u}\omega); (\cR^{r+u}\omega)_{r+u} ) = \bbE_{\omega} f(x_r(\omega); z_{r+u}).
    \end{equation*}
    Under WR or WOR sampling, conditional on $\cF_r$, the later indices follow the same uniform marginal distribution, 
    \begin{equation*}
    \bbE_{\omega}[f(x_r; z_{r+u}) \mid \cF_r] =  \bbE_{\omega}[f(x_r; z_{r+1}) \mid \cF_r].
    \end{equation*}
    Taking the full expectation gives $\bbE_{\omega} f(x_0 + Y_j(\cR \omega); z_m) = A_r$.
    By~\eqref{eq: traj}, we have
    \begin{equation*}
        X_{j+1}(\omega) - Y_{j+1}(\omega) = X_j(\cR \omega) - Y_j(\cR \omega) - G_j(\omega).
    \end{equation*}
    Hence, by the convexity, 
    \begin{equation*}
    \begin{aligned}
        \| X_{j+1}(\omega) - Y_{j+1}(\omega)\|_2^2 &= \| X_j(\cR \omega) - Y_j(\cR \omega) \|_2^2 + \|G_j(\omega)\|_2^2 \\ &\qquad - 2 \aver{ (x_0 + X_j(\cR \omega)) - (x_0 + Y_j(\cR \omega)) , G_j(\omega) } \\
        &\le  \| X_j(\cR \omega) - Y_j(\cR \omega) \|_2^2 + \|G_j(\omega)\|_2^2 \\&\qquad + 2 (  f(x_0 + Y_j(\cR \omega); z_{m})  -  f(x_0 + X_j(\cR \omega); z_{m})  ).
    \end{aligned}
    \end{equation*}
    Take the expectation over $\omega$, 
    \begin{equation*}
        \|X_{j+1} - Y_{j+1}\|_{\cH}^2 \le \|X_{j} - Y_{j}\|_{\cH}^2 + \|G_j\|_{\cH}^2 + 2(A_r - A_j).
    \end{equation*}
    Summing over $j$ and note that $Y_s = X_r$,
    \begin{equation*}
    \begin{aligned}
         \sum_{j=s}^{s + l - 1} \left( 2 A_j - \|G_j\|_{\cH}^2 \right) &\le 2l A_r + \|X_s - Y_s\|_{\cH}^2 - \|X_{s+l} - Y_{s+l}\|_{\cH}^2 \\
         &\le 2l A_r + \|X_s - X_r\|_{\cH}^2.
    \end{aligned}
    \end{equation*}
    Dividing the inequality above in half and applying~\eqref{eq: s-r} gives~\eqref{eq: traj comp}.
\end{proof}

\begin{corollary}\label{cor: Am}
    Under the same assumption in Lemma~\ref{lem: com run}, for a fixed $m\ge 2$, and any integer $\kappa\ge 2$, 
    \begin{equation*}
        A_{m-1} \le \frac{S_m}{m} + \frac{1}{2} \Big\lceil \frac{\log m}{\log \kappa}  \Big\rceil \left(\sup_{0\le j\le m-1} \|G_j\|_{\cH}^2  + 2(\kappa - 1) A_0 \right).
     \end{equation*}
\end{corollary}
\begin{proof}
    For a fixed $m$, let $M_{l} = \frac{1}{l}\sum_{j=m-l}^{m-1} A_j$, and $\overline{G}_m = \sup_{0\le j\le m-1} \|G_j\|_{\cH}^2$. Then set $s=m-l$ in~\eqref{eq: traj comp}, and for each $ 0\le r \le s - 1$, 
    \begin{equation*}
        M_l \le A_r + \frac{\overline{G}_m}{2} + \frac{2(s-r)A_0}{l}.
    \end{equation*}
    Average the above inequality over a short run of length $v$ with $r = s - v, \dots, s- 1$, where $1\le v \le s$, 
    \begin{equation}\label{eq: ave upp}
        M_l \le \frac{1}{v} \sum_{r = s-v}^{s-1} A_r + \frac{\overline{G}_m}{2} + \frac{(v+1) A_0 }{l}.
    \end{equation}
    Notice that 
    \begin{equation}\label{eq: ave id}
        (l+v) M_{l+v} = l M_l + \sum_{r = s-v}^{s-1} A_r. 
    \end{equation}
    Substitute~\eqref{eq: ave id} into~\eqref{eq: ave upp}, 
    \begin{equation}\label{eq: quota}
         M_l - M_{l+v} \le   \frac{v\overline{G}_m}{2(l+v)} + \frac{v(v+1)}{l(l+v) } A_0. 
    \end{equation}
    Set $l_p = \min\{\kappa^p, m\}$, and $M_{1} = A_{m-1}$, $M_{m} = \frac{1}{m}\sum_{j=0}^{m-1} A_j$. Then
    \begin{equation*}
        M_1 - M_m = \sum_{p=0}^{\lceil \log m/\log \kappa \rceil - 1} ( M_{l_p} - M_{l_{p+1}} ).
    \end{equation*}
    Apply~\eqref{eq: quota} to each term $M_{l_p} - M_{l_{p+1}}$, with $v = l_{p+1} - l_p \le l_p(\kappa - 1)$, we obtain $ \frac{v}{2(l+v)} < \frac{1}{2}$ and $ \frac{v(v+1)}{l(l+v) } \le \kappa - 1 $. Thus, rearrangement gives
    \begin{equation*}
        M_1 - M_m \le \lceil \log m/\log \kappa \rceil \left(\frac{\overline{G}_m}{2} + (\kappa - 1)A_0 \right).
    \end{equation*}
\end{proof}
Since $\kappa\ge 2$ is arbitrary and $\overline{G}_m\le (4 + \lceil \log_2 m \rceil)^2 A_0$ from Lemma~\ref{thm: grad}, we choose $\Gamma_1 = 4$ and let
\begin{equation*}
\begin{aligned}
    L_m &= 4 + \lceil \log_2 m \rceil, \\ 
    \Gamma_m &= 4 + \frac{1}{2} \min_{\substack{2 \le \kappa \le m\\ \kappa\in\bbN}} \Big\lceil \frac{\log m}{\log \kappa} \Big\rceil \left(  L_m^2 + 2(\kappa - 1) \right). 
\end{aligned}
\end{equation*}
Then we find that the balance $ L_m^2 \asymp \kappa$ happens when $\kappa\asymp (\log m)^2$. Therefore, $\Gamma_m\asymp (\log m)^3/\log\log m$. 
\begin{lemma}\label{lem: am bound}
Suppose $f(\,\cdot\,;z)$ is nonnegative, convex, and 1-smooth for every $z\in[n]$, and assume $x^{\ast}$ is the common zero for every $z\in[n]$. Let $\|x_0 - x^{\ast}\|_2=R$.
Then under either WR or WOR sampling, for every permitted $m\ge 1$, 
\begin{equation*}
    A_{m-1} \le \frac{\Gamma_{m} R^2}{2m}, \qquad \|G_{m-1}\|_{\cH}^2 \le \frac{L_m^2 R^2}{2m}.
\end{equation*}
\end{lemma}

\begin{proof}
    Let $s = 0,\dots, m-1$, define 
    \begin{equation*}
        \widetilde{A}_s = \bbE_{\omega}[f(x_s; z_{s+1})\mid \cF_s] = \begin{cases}
            F(x_s) \qquad &{\rm WR}, \\
            \frac{1}{n-s} \sum_{j\notin \{ z_1,\dots, z_s\} } f(x_s; j)\qquad &{\rm WOR}.
        \end{cases} 
    \end{equation*}
    Conditioned on $\cF_s$ and treating $s$ as the new starting index, by Lemma~\ref{thm: grad} and Corollary~\ref{cor: Am},
    \begin{equation*}
     \bbE_{\omega}[f(x_{m-1}; z_m)\mid \cF_s] \le \Gamma_{m-s} \widetilde{A}_s, \qquad  \bbE_{\omega}[\|\nabla f(x_{m-1}; z_m)\|_2^2\mid \cF_s] \le L_{m-s}^2 \widetilde{A}_s.
    \end{equation*}
    Taking expectations, and using $\bbE_{\omega}[\widetilde{A}_s] = A_s$, we find that
    \begin{equation*}
        A_{m-1} \le \Gamma_{m-s} A_s \le \Gamma_m A_s, \qquad  \|G_{m-1}\|_{\cH}^2 \le L_{m-s}^2 A_s \le L_m^2 A_s.
    \end{equation*}
    Sum over $s$, and use Corollary~\ref{cor: init dist}, 
    \begin{equation*}
        A_{m-1} \le \frac{\Gamma_m}{m} S_m \le \frac{\Gamma_m R^2}{2m},\qquad  \|G_{m-1}\|_{\cH}^2 \le \frac{L_m^2 R^2}{2m}.
    \end{equation*}
\end{proof}
In the Lemma~\ref{lem: am bound},  if we choose a suboptimal $\kappa = 2$, we can recover the upper bound $\cO((\log T)^3/T)$ as in~\citet{SchlisermanKoren2026} for WR sampling in the interpolation regime $\sigma_{\ast} = 0$. The next theorem shows this bound holds for both WR and WOR samplings.
\begin{theorem}\label{thm: no noise bd}
Suppose $f(\,\cdot\,;z)$ is nonnegative, convex, and 1-smooth for every $z\in[n]$, and assume $x^{\ast}$ is the common zero for every $z\in[n]$. Let $\|x_0 - x^{\ast}\|_2=R$.
For WR at every $T\ge 1$, and for WOR at every $1\le T \le n$, 
\begin{equation*}
    \bbE_{\omega}[F(x_T)] \le \frac{C  R^2 \ell_T^3}{T \lambda_T},\qquad    \bbE_{\omega}[\cG(x_{T})] \le \frac{C R^2  \ell_T^2}{T  },
\end{equation*}
where $C$ is an absolute constant, and $\ell_T, \lambda_T$ are defined in~\eqref{eq: log}. 
\end{theorem}
\begin{proof}
    The proof for WR sampling follows from Lemma~\ref{lem: am bound} directly with $m=T+1$, 
    \begin{equation*}
        \bbE_{\omega}[F(x_T)] = A_T \le \frac{\Gamma_{T+1}R^2}{2(T+1)},\qquad \bbE_{\omega}[\cG(x_{T})] = \|G_{T}\|_{\cH}^2 \le \frac{L_{T+1}^2 R^2}{2(T+1)}.  
    \end{equation*}
    For WOR sampling, we have 
    \begin{equation*}
    \begin{aligned}
        \bbE_{\omega}[F(x_T)] &= \frac{1}{n} \left( \sum_{j=1}^{T}\bbE[ f(x_T; z_j) ] + \sum_{l=T+1}^n \bbE[f(x_T; z_l)]   \right)\\
        &= \frac{T}{n} \bbE\left[ \frac{1}{T}\sum_{j=1}^{T} f(x_T; z_j) \right] + \frac{n-T}{n} A_T\\
        \bbE_{\omega}[\cG(x_T)] &= \frac{T}{n}\bbE\left[\frac{1}{T}\sum_{j=1}^T \|\nabla f(x_T; z_j)\|^2 \right] + \frac{n-T}{n} \|G_{T}\|_{\cH}^2 .
    \end{aligned}
    \end{equation*}
    To estimate $\bbE[ f(x_T; z_j) ]$, $1\le j\le T$, we consider the trajectory that skips the $j$-th index, denote the trajectory by $\{y_k^{(j)}\}_{k=0}^{T-1}$. Then
    \begin{equation*}
        y_{k}^{(j)} = x_k,\qquad 0\le k\le j-1,
    \end{equation*}
    and $x_{j} = x_{j-1} - \nabla f(x_{j-1}; z_j)$. For $k\ge j+1$, 
    \begin{equation*}
        x_{k} = x_{k-1} - \nabla f(x_{k-1}, z_k),\qquad y^{(j)}_{k-1} = y^{(j)}_{k-2} - \nabla f(y^{(j)}_{k-2}, z_{k}).
    \end{equation*}
    Using the non-expansiveness in Lemma~\ref{lem: nonex} repeatedly gives 
    \begin{equation*}
    \begin{aligned}
        \|x_T - y^{(j)}_{T-1}\|_2  &\le \|x_{j} - y^{(j)}_{j-1}\|_2 \\& = \|x_{j} - x_{j-1}\|_2 = \|\nabla f(x_{j-1}, z_j)\|_2.
    \end{aligned}
    \end{equation*}
    Sum over $1\le j\le T$, and apply Corollary~\ref{cor: init dist}, 
    \begin{equation*}
         \sum_{j=1}^T \|x_T - y^{(j)}_{T-1}\|^2_2  \le \sum_{j=1}^T \|\nabla f(x_{j-1}, z_j)\|_2^2 \le R^2. 
    \end{equation*}
    Moreover, since $z_j$ is not used in the trajectory of $\{y_{k}^{(j)}\}_{k=0}^{T-1}$, we have 
    \begin{equation*}
        \bbE_{\omega} [ f(y_{T-1}^{(j)}; z_j) ] = A_{T-1},\qquad \bbE_{\omega}[ \|\nabla f(y_{T-1}^{(j)}; z_j)\|_2^2] = \|G_{T-1}\|_{\cH}^2.
    \end{equation*}
By Lemma~\ref{lem: nonex} and 1-smoothness, 
\begin{equation*}
\begin{aligned}
   f(x_T; z_j)  -  f(y_{T-1}^{(j)}; z_j)  &\le  \aver{ \nabla f(y_{T-1}^{(j)}; z_j),  (x_T - y_{T-1}^{(j)}) } + \frac{1}{2} \|x_T - y_{T-1}^{(j)}\|_2^2\\
   &\le \sqrt{2 f(y_{T-1}^{(j)}; z_j) }\| x_T - y_{T-1}^{(j)} \| + \frac{1}{2} \|x_T - y_{T-1}^{(j)}\|_2^2 \\
   &\le f(y_{T-1}^{(j)}; z_j)  + \|x_T - y_{T-1}^{(j)}\|_2^2.
\end{aligned}
\end{equation*}
Sum over $j=1,\dots, T$, and take the expectation,
\begin{equation*}
\begin{aligned}
   \frac{1}{T}\sum_{j=1}^T \bbE_{\omega}[f(x_T, z_j)] &\le \frac{2}{T} \sum_{j=1}^T \bbE_{\omega}[f(y_{T-1}^{(j)}, z_j)]  + \frac{1}{T}\sum_{j=1}^T \bbE_{\omega} [ \|x_T - y_{T-1}^{(j)}\|_2^2 ] \\& \le  2A_{T-1} + \frac{R^2}{T}.
\end{aligned}
\end{equation*}
    Using the bound in Lemma~\ref{lem: am bound}, $A_{T-1}, A_T \le \Gamma_{T+1}R^2/(2T)$, 
    \begin{equation*}
    \begin{aligned}
      \bbE_{\omega}[F(x_T)] &\le \frac{T}{n} (  2A_{T-1} + \frac{R^2}{T} ) + \frac{n-T}{n} A_T  \\ &\le \frac{(\Gamma_{T+1} + 1)R^2}{T}.
    \end{aligned}
    \end{equation*}
Similarly, 1-smoothness implies 
\begin{equation*}
\begin{aligned}
  \|\nabla f(x_T; z_j) \|_2^2 &\le  2\|\nabla f(y_{T-1}^{(j)}; z_j)\|_2^2 +  2\|\nabla f(x_T; z_j) - \nabla f(y_{T-1}^{(j)}; z_j)\|^2_2 \\ & \le 2\|\nabla f(y_{T-1}^{(j)}; z_j)\|_2^2 + 2\|x_T - y_{T-1}^{(j)} \|_2^2.
\end{aligned}
\end{equation*}
Sum over $j=1,\dots, T$ and take expectation, 
\begin{equation*}
    \bbE_{\omega}\left[\frac{1}{T}\sum_{j=1}^T \|\nabla f(x_T; z_j)\|^2 \right] \le 2 \|G_{T-1}\|_{\cH}^2 + \frac{2 R^2}{T}.
\end{equation*}
Therefore, 
\begin{equation*}
\begin{aligned}
     \bbE_{\omega}[\cG(x_T)] &\le \frac{T}{n} \left( 2 \|G_{T-1}\|_{\cH}^2 + \frac{2 R^2}{T} \right) + \frac{n-T}{n} \|G_{T}\|_{\cH}^2 \\
     &\le \frac{(L_{T+1}^2 + 2) R^2}{T}.
\end{aligned}
\end{equation*}
\end{proof}
For the problem setting in Section~\ref{sec: pro}, after applying the above theorem to the normalized function $(f(\,\cdot\,; z) - f(x^{\ast}; z) )/\beta$, we can obtain the estimates
\begin{equation*}
    \bbE_{\omega}[F(x_T) - F(x^{\ast}) ] \le \frac{C \beta R^2 \ell_T^3}{T\lambda_T},\qquad    \bbE_{\omega}[\cG(x_{T})] \le \frac{C \beta^2 R^2 \ell_T^2 }{T},
\end{equation*}
which proves the improved upper bound $\cO(\beta R^2 (\log T)^3/(T\log\log T))$ for Corollary~\ref{thm: inter}. The polylogarithmic factors come from the dyadic partition in Lemma~\ref{thm: grad} and the trajectory comparison in Corollary~\ref{cor: Am}. To further reduce the factors, one possibility is to control gradient alignment across different scales in Lemma~\ref{thm: grad}. 

\section{Low-noise regime}\label{sec: noise}
In this section, we prove Theorem~\ref{thm: main} by extending the previous analysis to allow $f(\,\cdot\,; z)$ to have different minimizers while their mean $F$ has a minimizer $x^{\ast}$. Fix $\eta \in (0, 2\beta^{-1})$, and let 
\begin{equation*}
\widetilde{f}(x; z) := \eta f(x; z),\qquad \alpha := \eta\beta \in (0, 2),\qquad \delta := \min\{1, 2-\alpha\},\qquad \sigma^2 := \eta^2 \sigma_{\ast}^2, 
\end{equation*}
We also denote $ \widetilde{F}(x) := \bbE_{z\sim \cZ}[\widetilde{f}(x; z)]$ and $\widetilde{\cG}(x):=  \bbE_{z\sim \cZ}[\|\nabla \widetilde{f}(x; z)\|_2^2]$. Let
\begin{equation*}
   \qquad \varphi(x; z):= \widetilde{f}(x; z) - \widetilde{f}(x^{\ast}; z) - \aver{\zeta_{z}, x-x^{\ast}},\qquad \zeta_z = \nabla \widetilde{f}(x^{\ast}; z).  
\end{equation*}
Then $\varphi(\;\cdot\; ; z)$ is non-negative, convex, $\alpha$-smooth, with $\varphi(x^{\ast}; z) = 0$, and $\nabla \varphi(x^{\ast}; z) = 0$. Moreover, by the optimality of $x^{\ast}$, we have $\bbE_{z\sim\cZ} [\zeta_z] = 0$, and  
\begin{equation*}
\bbE_{z\sim\cZ}[\varphi(x; z)] = \widetilde{F}(x) - \widetilde{F}(x^{\ast}). 
\end{equation*}
The SGD update becomes 
\begin{equation*}
    x_{t} = x_{t-1} - \nabla \varphi(x_{t-1}; z_t) - \zeta_{z_t},\qquad t = 1,\dots, T.
\end{equation*}

\subsection{Preliminary estimates}
By slight modification of Lemma~\ref{lem: nonex}, the non-expansiveness can be shown for any non-negative, convex, $\alpha$-smooth function $h$ with $\alpha\in(0,2)$:
\begin{equation}\label{eq: nonex noise}
    \begin{aligned}
    \|(u - \nabla h(u)) - (v - \nabla h(v))\|_2^2 + \frac{2-\alpha}{\alpha}\|\nabla h(u) - \nabla h(v)\|_2^2 &\le \|u - v\|_2^2,\\
    \|\nabla h(u)\|_2^2 &\le 2 \alpha h(u).
    \end{aligned}
\end{equation}
The first is equivalent to the co-coercivity $\aver{\nabla h(u) - \nabla h(v), u - v} \ge  \alpha^{-1} \|\nabla h(u) -\nabla h(v)\|_2^2$. The second applies the same derivation~\eqref{eq: low} for $h(u - \alpha^{-1}\nabla h(u))$. 
\begin{lemma}\label{lem: stat}
Let $\zeta_1,\dots, \zeta_n$ be fixed vectors and 
\begin{equation*}
    \mu = \frac{1}{n} \sum_{i=1}^n \zeta_i,\qquad \nu^2 = \frac{1}{n} \sum_{i=1}^n \|\zeta_i\|_2^2. 
\end{equation*}
Let $\omega = (z_1,\dots, z_n)$ be drawn from uniform WOR samplings and define
\begin{equation*}
    \mu_r(\omega) = \frac{1}{n-r}\sum_{i=r+1}^n\zeta_{z_i}. 
\end{equation*}
Then for WOR sampling with $n\ge 2$ and $0\le r\le n-1$, 
\begin{equation*}
    \bbE_{\omega} [\|\mu_r(\omega) - \mu\|_2^2] = \frac{r}{(n-1)(n-r)} (\nu^2 - \|\mu\|^2) \le \frac{1}{n-r} (\nu^2 - \|\mu\|^2) . 
\end{equation*}
\end{lemma}
\begin{proof}
First, we notice that if $i\neq j$, 
\begin{equation*}
    \bbE_{\omega}[\aver{\zeta_{z_i} - \mu, \zeta_{z_j} - \mu}] = - \frac{n (\nu^2 - \|\mu\|_2^2) }{n(n-1)} = -\frac{ \nu^2 - \|\mu\|_2^2 }{n-1}  .
\end{equation*}
Therefore,
\begin{equation*}
    \bbE_{\omega}[\|\sum_{j=1}^r (\zeta_{z_j} - \mu)\|^2_2] = \left( r -  \frac{r(r-1)}{n-1} \right) (\nu^2 - \|\mu\|_2^2 ) = \frac{r(n-r)}{n-1} (\nu^2 - \|\mu\|_2^2 ).
\end{equation*}
Since $\mu_r - \mu = -(n-r)^{-1} \sum_{j=1}^r (\zeta_{z_j} - \mu)$, 
\begin{equation*}
    \bbE_{\omega}[\|\mu_r - \mu\|_2^2] = \frac{1}{(n-r)^2} \frac{r(n-r)}{n-1} (\nu^2 - \|\mu\|_2^2 ) = \frac{r}{(n-1)(n-r)} (\nu^2 - \|\mu\|_2^2 ).
\end{equation*}
Since $r\le n-1$, it leads to a simpler bound $\frac{1}{n-r} (\nu^2 - \|\mu\|_2^2 ) $.
\end{proof}
The following estimates are standard, while the constants are not intentionally optimized. The constants for WR sampling can be significantly smaller.
\begin{lemma}\label{lem: dist noise}
    For both WR and WOR sampling, and every permitted batch length $m$, 
    \begin{equation*}
        \bbE_{\omega}[\|x_m - x^{\ast}\|_2^2] + \delta \sum_{j=0}^{m-1} \bbE_{\omega}[\varphi(x_j; z_{j+1})] \le 2R^2 + \frac{26 m \sigma^2}{\delta}. 
    \end{equation*}
\end{lemma}

\begin{proof}
    Let $p_{j-1} = x_{j-1} - x^{\ast}$, and $q_j = \nabla \varphi(x_{j-1}, z_j)$, then $p_{j} = p_{j-1} - q_j - \zeta_{z_j}$, and
    \begin{equation*}
        0 = \varphi(x^{\ast}; z_j) \ge \varphi(x_{j-1}; z_j) - \aver{q_j, p_{j-1}} + \frac{1}{2\alpha} \|q_{j}\|_2^2. 
    \end{equation*}
    Hence, similar to Corollary~\ref{cor: init dist},
    \begin{equation*}
        \|p_{j}\|_2^2 - \|p_{j-1}\|_2^2 \le -2 \varphi(x_{j-1}; z_j) + (1 - \alpha^{-1})\|q_{j}\|_2^2 + 2\aver{q_j, \zeta_{z_j}} + \|\zeta_{z_j}\|_2^2 - 2\aver{p_{j-1}, \zeta_{z_j}}.
    \end{equation*}
    Denote $z_{+} = \max(z, 0)$, then $\delta = 1 - (\alpha - 1)_{+}$.  By~\eqref{eq: nonex noise},
    \begin{equation*}
         (1 - \alpha^{-1})\|q_{j}\|_2^2 \le 2(\alpha - 1)_{+} \varphi(x_{j-1}; z_j),
    \end{equation*}
    and
    \begin{equation*}
    2\aver{q_j, \zeta_{z_j}} \le \frac{\delta}{2\alpha} \|q_{j}\|_2^2 + \frac{2\alpha}{\delta} \|\zeta_{z_j}\|_2^2 \le \delta \varphi(x_{j-1}; z_j) + \frac{2\alpha}{\delta} \|\zeta_{z_j}\|_2^2 .
    \end{equation*}
    Combining the above gives 
    \begin{equation}\label{eq: tele}
         \|p_{j}\|_2^2 - \|p_{j-1}\|_2^2 \le -\delta \varphi(x_{j-1}; z_j) + \frac{2\alpha + \delta}{\delta} \|\zeta_{z_j}\|_2^2 - 2\aver{p_{j-1}, \zeta_{z_j}}.
    \end{equation}
    Under WR sampling, we take the expectation of the above, then $\bbE_{\omega}[\aver{p_{j-1}, \zeta_{z_j}}] = 0$, and  
    \begin{equation*}
        \bbE_{\omega}[\|p_{j}\|_2^2] - \bbE_{\omega}[\|p_{j-1}\|_2^2] + \delta \bbE_{\omega}[\varphi(x_{j-1}; z_j) ] \le \frac{2\alpha + \delta}{\delta} \bbE_{\omega}[\|\zeta_{z_j}\|_2^2]. 
    \end{equation*}
    Sum over $j=1,\dots, m$, and use the bounds $\alpha < 2$, $\delta \le 1$, 
    \begin{equation*}
    \begin{aligned}
          \bbE_{\omega}[\|x_m - x^{\ast}\|_2^2] - \bbE_{\omega}[\|x_0 -x^{\ast}\|_2^2] + \delta \sum_{j=1}^m \bbE_{\omega}[\varphi(x_{j-1}; z_j) ] & \le \frac{2\alpha + \delta}{\delta} \sum_{j=1}^m \bbE_{\omega}[\|\zeta_{z_j}\|_2^2] \\
          &\le \frac{5 m \sigma^2}{\delta} .
    \end{aligned}
    \end{equation*}
    Under WOR sampling, we assume $n\ge 2$ to avoid triviality and fix $m\le n$ and let $\cD_m = \sup_{0\le r \le m} (\bbE_{\omega}[\|x_r - x^{\ast}\|^2])^{1/2}$. Conditioning on the first $j-1$ updates and apply Lemma~\ref{lem: stat} with $\mu = 0$ and $\nu^2 = \sigma^2$,  
    \begin{equation*}
        \left| \bbE_{\omega}[\aver{\zeta_{z_{j}}, p_{j-1}}] \right| =   \left| \bbE_{\omega}[\aver{\mu_{j-1}, p_{j-1}}] \right|\le \sigma \cD_m \sqrt{\frac{j-1}{(n-1)(n-(j-1))}} \le \frac{\sigma \cD_m}{\sqrt{ n - (j-1)}}.  
    \end{equation*}
    Summing~\eqref{eq: tele} over $j=1,\dots, l$ for some $l\le m$, 
    \begin{equation}\label{eq: pl}
    \begin{aligned}
    \bbE_{\omega}[\|p_{l}\|_2^2] + \delta \sum_{j=1}^l \bbE_{\omega}[\varphi(x_{j-1}; z_j)] &\le R^2 + \frac{2\alpha + \delta}{\delta} m \sigma^2 + 2\sigma \cD_m \sum_{j=1}^l (n - (j-1))^{-1/2} \\
    &\le R^2 + \frac{2\alpha + \delta}{\delta} m \sigma^2 + 2\sigma \cD_m \sum_{j=1}^l (l - (j-1))^{-1/2} \\
    &\le R^2 + \frac{2\alpha + \delta}{\delta} m \sigma^2 + 4\sigma \cD_m \sqrt{m}.
    \end{aligned}
    \end{equation}
    Taking the supremum over $ \bbE_{\omega}[\|p_{l}\|_2^2] $ over $0\le l \le m$ on the left-hand side, and then applying the AM-GM inequality, we obtain
    \begin{equation*}
    \begin{aligned}
        \cD_m^2 &\le R^2+ \frac{2\alpha + \delta}{\delta} m \sigma^2 + 4\sigma \cD_m \sqrt{m} \\ & \le R^2 + \frac{2\alpha + \delta}{\delta} m \sigma^2 + \frac{1}{2} ( \cD_m^2 + 16 m \sigma^2 ) .
    \end{aligned}
    \end{equation*}
    Thus, $\cD_m \le 2R^2 + \left( \frac{2(2\alpha + \delta)}{\delta} + 16 \right) m \sigma^2 \le 2R^2 + \frac{26 m \sigma^2}{\delta}$. Applying this to the right-hand side of~\eqref{eq: pl} gives the desired bound.
\end{proof}

\begin{corollary}\label{cor: grad sum}
    For both WR and WOR sampling, and every permitted batch length $m$, 
\begin{equation*}
    \sum_{j=0}^{m-1} \bbE_{\omega}[\|\nabla \widetilde{f}(x_j; z_{j+1})\|_2^2] \le \frac{8 \alpha R^2}{\delta} + \left(\frac{104\alpha}{\delta^2} + 2\right) m\sigma^2.
\end{equation*}
\end{corollary}
\begin{proof}
    Apply the estimate~\eqref{eq: nonex noise}, 
    \begin{equation*}
    \begin{aligned}
     \|\nabla \widetilde{f}(x_j; z_{j+1})\|_2^2 &=  \|\nabla \varphi(x_j; z_{j+1}) + \zeta_{z_{j+1}}\|_2^2 \\ &\le 2 \|\nabla \varphi(x_j; z_{j+1})\|_2^2 + 2 \|\zeta_{z_{j+1}}\|_2^2 \\&\le 4 \alpha \varphi(x_j; z_{j+1}) + 2 \|\zeta_{z_{j+1}}\|_2^2.
    \end{aligned}
    \end{equation*}
    Take summation over $j$ and expectation over $\omega$, 
    \begin{equation*}
          \sum_{j=0}^{m-1} \bbE_{\omega}[\|\nabla \widetilde{f}(x_j; z_{j+1})\|_2^2] \le 4\alpha  \sum_{j=0}^{m-1} \bbE_{\omega}[\varphi(x_j; z_{j+1})]  + 2m \sigma^2.
    \end{equation*}
    Using Lemma~\ref{lem: dist noise}, 
    \begin{equation*}
          \sum_{j=0}^{m-1} \bbE_{\omega}[\|\nabla \widetilde{f}(x_j; z_{j+1})\|_2^2] \le 4\alpha \frac{2R^2 + 26m\sigma^2/\delta}{\delta} + 2m\sigma^2 =  \frac{8 \alpha R^2}{\delta} + \left(\frac{104\alpha}{\delta^2} + 2\right) m\sigma^2.
    \end{equation*}
\end{proof}

The following estimate is analogous to Lemma~\ref{lem: init} and Corollary~\ref{cor: init dist}.
\begin{lemma}\label{lem: n diff}
    Let $h_1,\dots, h_n$ be non-negative, convex, $\alpha$-smooth functions. Let $\zeta_1,\dots, \zeta_n$ be fixed vectors with
    \begin{equation*}
        \frac{1}{n}\sum_{i=1}^n \zeta_i = \mu,\qquad \frac{1}{n} \sum_{i=1}^n \|\zeta_i\|_2^2 = \nu^2.
    \end{equation*}
    Fix a deterministic $u_0$ and consider the iteration $u_{j} = u_{j-1} - \nabla h_{z_j}(u_{j-1}) - \zeta_{z_j}$. Then under WR or WOR samplings of $\omega$ with permitted length $m\ge 1$, and any $0\le s\le m$, 
    \begin{equation}\label{eq: u-u0}
    \begin{aligned}
        \bbE_{\omega} [\|u_s - u_0\|_2^2 ] + \delta \sum_{j=1}^s \bbE_{\omega}[h_{z_j}(u_{j-1})] &\le s K_m,\\
        \sum_{j=1}^s \bbE_{\omega} [\|u_j - u_{j-1}\|_2^2 ] &\le s \left(\frac{4\alpha K_m}{\delta} + 2 \nu^2\right), 
    \end{aligned}
    \end{equation}
    where 
    \begin{equation}\label{eq: km}
        K_m = \frac{16}{n\delta} \sum_{i=1}^n h_{i}(u_0)  + \left(34 + \frac{8\alpha}{\delta}\right) \nu^2  + 8m \|\mu\|_2^2.  
    \end{equation}
\end{lemma}
\begin{proof}
If $s=0$, the estimates are trivial. We now consider $1\le s\le m$. 
    Let $p_{j-1} = u_{j-1} - u_0$, $q_j = \nabla h_{z_j}(u_{j-1})$ and $r_j = \nabla h_{z_j}(u_0)$, then $p_j = p_{j-1} - q_j - \zeta_{z_j}$, and
    \begin{equation*}
        h_{z_j}(u_0) \ge h_{z_j}(u_{j-1}) - \aver{q_{j}, p_{j-1}} + \frac{\|q_j -r_j\|_2^2}{2\alpha}.
    \end{equation*}
    Then
    \begin{equation*}
    \begin{aligned}
        \|p_{j-1} - q_j - \zeta_{z_j}\|_2^2 - \|p_{j-1}\|_2^2 & \le 2 h_{z_j}(u_0) - 2 h_{z_j}(u_{j-1}) + (1-\alpha^{-1})\|q_j\|_2^2 + 2 \aver{q_j, \frac{r_j}{\alpha}  + \zeta_{z_j}} \\& \qquad - \frac{\|r_j\|_2^2}{\alpha} + \|\zeta_{z_j}\|_2^2 - 2 \aver{\zeta_{z_j}, p_{j-1}}.
    \end{aligned}
    \end{equation*}
    Similar to Lemma~\ref{lem: dist noise}, we have 
    \begin{equation*}
       -2 h_{z_j}(u_{j-1}) + (1 - \alpha^{-1}) \|q_j\|_2^2 \le -2 h_{z_j}(u_{j-1}) + 2 (\alpha - 1)_{+} h_{z_j}(u_{j-1}) = -2\delta h_{z_j}(u_{j-1}).
    \end{equation*}
    Moreover, $\|r_j\|_2^2 \le 2 \alpha h_{z_j}(u_0)$, $\|q_j\|_2^2 \le 2 \alpha h_{z_j}(u_{j-1})$ and
    \begin{equation*}
        -\frac{\delta}{2\alpha} \|q_j\|_2^2 + 2 \aver{q_j, \frac{r_j}{\alpha} + \zeta_{z_j}} \le \frac{2\alpha}{\delta} \|\frac{r_j}{\alpha} + \zeta_{z_j}\|_2^2 \le \frac{4}{\alpha \delta} \|r_j\|_2^2 + \frac{4\alpha}{\delta} \|\zeta_{z_j}\|_2^2.
    \end{equation*}
    Combining the above estimates, 
    \begin{equation*}
        \|p_{j-1} - q_j - \zeta_{z_j}\|_2^2 - \|p_{j-1}\|_2^2 + \delta h_{z_j}(u_{j-1}) \le  \frac{8 h_{z_j}(u_0)}{\delta} + \left(1 + \frac{4\alpha}{\delta}\right) \|\zeta_{z_j}\|_2^2 -2\aver{\zeta_{z_j}, p_{j-1}}.
    \end{equation*}
    Summing over $j=1,\cdots, l$ with $l\le s$ and take expectation, set $C_0 = \frac{8}{n \delta} \sum_{i=1}^n h_i(u_0) +  \left(1 + \frac{4\alpha}{\delta}\right) \nu^2 $. Then
    \begin{equation}\label{eq: ul}
        \bbE_{\omega}[\|u_{l} - u_0\|_2^2] + \delta \sum_{j=1}^l \bbE_{\omega}[h_{z_j}(u_{j-1})] \le l C_0 - 2 \sum_{j=1}^l \bbE_{\omega}[\aver{\zeta_{z_j}, p_{j-1}}],
    \end{equation}
    Denote $\cD_s = \sup_{1\le j\le s} \bbE_{\omega}[\|p_{j}\|_2^2]^{1/2}/\sqrt{j}$. 
    Under WR sampling, for $1\le j\le s$, $$ \left| \bbE_{\omega}[\aver{\zeta_{z_j}, p_{j-1}}] \right| \le \|\mu\| \sqrt{j} \cD_s. $$ Under WOR sampling, by Lemma~\ref{lem: stat}, 
    $$\left| \bbE_{\omega}[\aver{\zeta_{z_j}, p_{j-1}}] \right| \le   \left(\|\mu\| + \nu (n - (j-1))^{-1/2} \right) \sqrt{j} \cD_s. $$
    Thus, for either case, we estimate the sum over $j=1,\dots, l$,
    \begin{equation*}
        - 2 \sum_{j=1}^l \bbE_{\omega}[\aver{\zeta_{z_j}, p_{j-1}}] \le 
        l ( 2 \sqrt{s} \|\mu\| + 4\nu) \cD_s,
    \end{equation*}
    where we have used 
    \begin{equation*}
      \sum_{j=1}^l (n - (j-1))^{-1/2} \sqrt{j} \le \sqrt{l} \int_0^l (l-x)^{-1/2} dx = 2l. 
    \end{equation*}
    Applying the above bounds to~\eqref{eq: ul}, we obtain for $1\le l\le s$,
    \begin{equation*}
        \bbE_{\omega}[\|u_{l} - u_0\|_2^2] \le l( C_0 + (2\sqrt{s} \|\mu\| + 4 \nu ) \cD_s ).
    \end{equation*}
    Hence, dividing the above by $l$ on both sides and maximizing on the left-hand side over $l$, we have $$\cD_s^2 \le C_0 + \frac{1}{2}(2\sqrt{s} \|\mu\| + 4 \nu )^2 + \frac{\cD_s^2}{2}.$$
    Thus, $\cD_s^2 \le 2 C_0 + 8 s \|\mu\|_2^2 + 32 \nu^2$, which is also the upper bound of~\eqref{eq: ul}. Taking $l=s$ gives the first bound in~\eqref {eq: u-u0}. 

    The second bound of~\eqref{eq: u-u0} follows from 
    \begin{equation*}
    \begin{aligned}
            \|u_j - u_{j-1}\|_2^2 = \|\nabla h_{z_j}(u_{j-1}) + \zeta_{z_j}\|_2^2 &\le 2 \|\nabla h_{z_j}(u_{j-1}) \|_2^2 + 2 \|\zeta_{z_j}\|_2^2 \\&\le 4\alpha h_{z_j}(u_{j-1}) + 2 \|\zeta_{z_j}\|_2^2.
    \end{aligned}
    \end{equation*}
    Summing over $j$ and applying the first bound leads to the desired result.
\end{proof}

\subsection{Gradient estimate}
We apply the invariant sampling space introduced in Section~\ref{sec: inv} to the sequence $\{u_k\}_{0\le k \le m}$. Let 
\begin{equation*}
    \psi(x; i) := h_i(x) + \aver{\zeta_i, x - x^{\ast}}.
\end{equation*}
The update is $u_{j} = u_{j-1} - \nabla \psi(u_{j-1}; z_j)$. Reusing the definition~\ref{eq: traj}, starting from $X_0 = 0$, and for $0\le k < m$,
\begin{equation*}
\begin{aligned}
    G_k(\omega) &= \nabla h_{z_m}(u_0 + X_k(\cR\omega)) + \zeta_{z_m} \\
    X_{k+1} &= X_k(\cR\omega) - G_k(\omega). 
\end{aligned}
\end{equation*}
In addition, we let $A_k := \bbE_{\omega} [\psi(u_0 + X_k(\cR\omega); z_m)]$. The identities from Lemma~\ref{lem: bridge} and Corollary~\ref{cor: h norm} give
\begin{equation*}
    \begin{aligned}
        u_k(\cR^k \omega) &= u_0 + X_k(\omega),\qquad &0\le k \le m, \\
        G_k(\omega) &= \nabla \psi(u_k(\cR^{k+1}\omega), (\cR^{k+1}\omega)_{k+1}),\qquad & 0\le k < m.
    \end{aligned}
\end{equation*}
Therefore, 
    \begin{equation*}
    \begin{aligned}
            \|X_k\|_{\cH}^2 &= \bbE_{\omega}[\|u_{k}(\omega) - u_0\|_2^2], \\
            \|G_k\|_{\cH}^2 &= \bbE_{\omega}[\|\nabla h_{z_{k+1}} (u_{k}(\omega)) + \zeta_{z_{k+1}}\|_2^2],\\
            A_k &= \bbE_{\omega}[\psi(u_k; z_{k+1})].
    \end{aligned}
    \end{equation*}
Next, we provide an analog to Lemma~\ref{thm: grad}.
\begin{lemma}\label{lem: grad noise}
    Under the same assumption of Lemma~\ref{lem: n diff},
    \begin{equation*}
     \|G_{m-1}\|_{\cH}^2 = \bbE_{\omega} [\|\nabla h_{z_m}(u_{m-1}) + \zeta_{z_m}\|_2^2] \le\frac{(32 + \lfloor \log_2 m \rfloor^2)\alpha K_m}{2\delta} + 8\nu^2. 
    \end{equation*}
\end{lemma}
\begin{proof}
Fix a permitted length $m$. By Lemma~\ref{lem: n diff}, we have 
    \begin{equation*}
        \sum_{k=0}^{m-1} \|G_k\|_{\cH}^2 \le m \left(\frac{4\alpha K_m}{\delta} + 2\nu^2\right).
    \end{equation*}
    Similar to Lemma~\ref{lem: diff}, we apply the relation~\eqref{eq: nonex noise},
\begin{equation*}
\begin{aligned}
 \|X_{k+1} - X_{k-s+1}\|_{\cH}^2 + \frac{2-\alpha}{\alpha}\| G_{k} - G_{k-s} \|_{\cH}^2 &\le \bbE_{\omega}[\| X_{k}(\cR\omega) - X_{k - s}(\cR\omega)\|_2^2]\\&= \|X_{k} -X_{k - s} \|_{\cH}^2.
\end{aligned}
\end{equation*}
Thus, use the first estimate of Lemma~\ref{lem: n diff},
\begin{equation*}
    \sum_{k=s}^{m-1} \| G_{k} - G_{k-s} \|_{\cH}^2 \le  \frac{\alpha}{2-\alpha} \|X_s\|_{\cH}^2 \le \frac{s \alpha K_m}{\delta}.
\end{equation*}
Follow the same computation in Lemma~\ref{thm: grad}. Let $U_{l} = \frac{1}{l}\sum_{k=m-l}^{m-1}G_k$ and set $J= \lfloor \log_2 m \rfloor$. Then
\begin{equation*}
\begin{aligned}
    \|G_{m-1}\|_{\cH} &\le \|U_{2^J}\|_{\cH} + \sum_{j=0}^{J-1} \|U_{2^j} - U_{2^{j+1}}\|_{\cH} \\
    &\le \|U_{2^J}\|_{\cH} + J \sqrt{\frac{\alpha K_m}{4\delta}  } \\
    &\le \sqrt{\frac{m}{2^J}   \left(\frac{4\alpha K_m}{\delta} + 2\nu^2\right) } +  J \sqrt{\frac{\alpha K_m}{4\delta}  }.
\end{aligned}
\end{equation*}
Therefore, using $m < 2^{J+1}$, 
\begin{equation*}
      \|G_{m-1}\|_{\cH}^2 \le 2   \left(\frac{8\alpha K_m}{\delta} + 4\nu^2 \right) + 2 J^2 \frac{\alpha K_m}{4\delta}  = \frac{(32 + J^2)\alpha K_m}{2\delta} + 8\nu^2 .
\end{equation*}
\end{proof}

\subsection{Trajectory comparison}
We now use the same trajectory comparison as in Section~\ref{sec: tr comp}. Let $\Theta_1 = 2$ and 
\begin{equation*}
\begin{aligned}
L_m &:= 4 + \lfloor \log_2 m \rfloor,\\
    \Theta_m &:= 2 + \frac{1}{2} \min_{\substack{2\le \kappa < m \\ \kappa\in\bbN}} \Big\lceil \frac{\log m}{\log \kappa}\Big\rceil \left(  \alpha L_m^2 + 8 + \frac{\delta}{2}(\kappa - 1)  \right). 
\end{aligned}
\end{equation*}
By taking $\kappa \asymp (\log m)^2$, we have $\Theta_m = \cO((\log m)^3/(m\log\log m))$.
\begin{lemma}\label{lem: n am}
    For every permitted length $m\ge 1$,
    \begin{equation*}
        A_{m-1} \le \aver{\mu, u_0 - x^{\ast}} + \frac{\Theta_m}{\delta} K_m.
    \end{equation*}
\end{lemma}
\begin{proof}
    By Lemma~\ref{lem: n diff}, we have 
    \begin{equation}\label{eq: aver}
        \frac{1}{m}\sum_{j=0}^{m-1} \bbE_{\omega}[h_{z_{j+1}}(u_j)] \le \frac{K_m}{\delta},\qquad \|X_j\|_{\cH}^2 \le j K_m. 
    \end{equation}
    Notice that 
    \begin{equation*}
        \begin{aligned}
        A_j &= \bbE_{\omega} [\psi(u_j; z_{j+1})] \\& = \bbE_{\omega} [h_{z_{j+1}}(u_j)] + \bbE_{\omega}[\aver{\zeta_{z_{j+1}}, u_j - u_0}] + \bbE_{\omega}[\aver{\zeta_{z_{j+1}}, u_0 - x^{\ast}}].
        \end{aligned}
    \end{equation*}
    The last term is $\aver{\mu, u_0 - x^{\ast}}$. 
    Under WR, we have 
    \begin{equation*}
        \bbE_{\omega}[\aver{\zeta_{z_{j+1}}, u_j - u_0}] \le \|\mu\|_2 \sqrt{j K_m} .
    \end{equation*}
    Under WOR, we apply Lemma~\ref{lem: stat}, 
        \begin{equation*}
        \bbE_{\omega}[\aver{\zeta_{z_{j+1}}, u_j - u_0}] \le \left( \|\mu\|_2  + \nu (n-j)^{-1/2} \right) \sqrt{j K_m} .
    \end{equation*}
    Summing over $j$, and use the definition of $K_m$ in~\eqref{eq: km},  
    \begin{equation*}
    \begin{aligned}
            \frac{1}{m}\sum_{j=0}^{m-1} A_j &\le \aver{\mu, u_0 - x^{\ast}} + \frac{K_m}{\delta} + \left(\sqrt{m} \|\mu\|_2 + 2 \nu \right) \sqrt{K_m} \\
            &\le \aver{\mu, u_0 - x^{\ast}} + \frac{2 K_m}{\delta}. 
    \end{aligned}
    \end{equation*}
    By Lemma~\ref{lem: grad noise}, and using $ (32 + \lfloor \log_2 m \rfloor^2) \le 2 ( 4 + \lfloor \log_2 m \rfloor)^2$, we obtain
    \begin{equation*}
        \sup_{0\le j < m} \|G_{j}\|_{\cH}^2 \le \frac{(32 + \lfloor \log_2 m \rfloor^2)\alpha K_m}{2\delta} + 8\nu^2 \le \frac{\alpha L_m^2}{\delta} K_m + 8 \nu^2.
    \end{equation*}
    We emphasize that Lemma~\ref{lem: com run} still applies. It does not depend on $\psi$ being non-negative. The needed non-expansiveness comes from~\eqref{eq: nonex noise}. Then, for any $0\le r \le s$ and $s + l\le m$, 
\begin{equation}
\begin{aligned}
    \sum_{j=s}^{s + l - 1} \left( A_j - \frac{1}{2} \|G_j\|_{\cH}^2 \right) &\le l A
    _r + \frac{1}{2}\|X_{s} - X_r\|_{\cH}^2 \\& \le l A
    _r + \frac{1}{2} \|X_{s-r}\|_{\cH}^2 \le l A_r + \frac{s-r}{2} K_m.
\end{aligned}
\end{equation}
Then repeating the calculations in Corollary~\ref{cor: Am}, with $A_0$ replaced by $K_m/4$, the following holds for any integer $\kappa \ge 2$, 
\begin{equation*}
\begin{aligned}
    A_{m-1} & \le  \aver{\mu, u_0 - x^{\ast}} + \frac{2 K_m}{\delta} + \frac{1}{2} \Big\lceil \frac{\log m}{\log \kappa} \Big\rceil \left( \sup_{0\le j\le m-1} \|G_j\|_{\cH}^2 + \frac{K_m}{2} (\kappa - 1)    \right)  \\
    & \le \aver{\mu, u_0 - x^{\ast}} + \frac{2 K_m}{\delta} + \frac{1}{2} \Big\lceil \frac{\log m}{\log \kappa} \Big\rceil \left( \frac{\alpha L_m^2}{\delta} K_m + 8 \nu^2 + \frac{K_m}{2} (\kappa - 1)    \right) \\
    &\le \aver{\mu, u_0 - x^{\ast}} + \left[ 2  + \frac{1}{2} \Big\lceil \frac{\log m}{\log \kappa} \Big\rceil \left( {\alpha L_m^2} + 8 + \frac{\delta}{2} (\kappa - 1)    \right)  \right]  \frac{K_m}{\delta},
\end{aligned}
\end{equation*}
where we have used $K_m \ge \delta \nu^2$. Optimizing over $\kappa$ gives the needed bound. The edge case $m=1$ directly follows from~\eqref{eq: aver}.
\end{proof}

\subsection{Proof of Theorem~\ref{thm: main}}
We return to the original trajectory $\{ x_j\}_{0\le j\le T}$ by taking $$\psi(x; z) = \widetilde{f}(x; z) - \widetilde{f}(x^{\ast}; z) = \varphi(x; z) + \aver{\zeta_{z}, x - x^{\ast}}.$$
Then $A_k = \bbE_{\omega}[\psi(x_k; z_{k+1})]$ and $\|G_k\|_{\cH}^2 = \bbE_{\omega}[\|\nabla \widetilde{f}(x_k; z_{k+1})\|_2^2]$.

\begin{lemma}\label{lem: n bd}
    For a permitted length $m\ge 1$, 
    \begin{equation*}
        \begin{aligned}
            A_{m-1} &\le \Theta_m \left( \frac{65 R^2}{2 m\delta^3} + \frac{961 \sigma^2}{2\delta^4} \right),  \\
            \|G_{m-1}\|_{\cH}^2 &\le \frac{32\alpha L_m^2 R^2}{m\delta^3} + \left(\frac{474\alpha L_m^2}{\delta^4} + 8\right) \sigma^2. 
        \end{aligned}
    \end{equation*}
\end{lemma}
\begin{proof}
    For each $0\le r < m$, we condition on $\cF_r$ and restart the trajectory at $x_r$. 
    Let $\mu_r$ and $\nu_r^2$ be the mean and second moment of $\zeta_i$ for the remaining unused $(m-r)$ indices. Under WR, $\mu_r = 0$ and $\nu_r^2 = \sigma^2$. Under WOR, $\bbE_{\omega}[\nu_r^2] = \sigma^2$ and $\bbE_{\omega}[\|\mu_r\|_2^2]\le \sigma^2/(m-r)$ by Lemma~\ref{lem: stat}. On the restarted trajectory, we only need to replace $K_m$ in Lemma~\ref{lem: n diff} by
\begin{equation*}
   K_{m-r, r} = \frac{16}{\delta} \mathbb E_{\omega}[\varphi(x_r;z_{r+1})\mid\mathcal F_r]  + \left(34 + \frac{8\alpha}{\delta}\right) \nu_r^2  + 8(m-r) \|\mu_r\|_2^2.
\end{equation*}
    Then apply the conditional expectation to Lemma~\ref{lem: n am}, 
    \begin{equation*}
     \bbE_{\omega}[\psi(x_{m-1}; z_m) \mid \cF_r] \le \aver{\mu_r, x_r - x^{\ast}} + \frac{\Theta_{m-r}}{\delta} K_{m-r, r}.
    \end{equation*}
    Take the full expectation recovers $A_{m-1}$ on the left-hand side, and use $\Theta_{m-r} \le \Theta_m$, 
    \begin{equation*}
        A_{m-1} \le \bbE_{\omega}[\aver{\mu_r, x_r - x^{\ast}}] + \frac{\Theta_{m}}{\delta} \bbE_{\omega}[K_{m-r, r}].
    \end{equation*}
    Similarly, by Lemma~\ref{lem: grad noise} and $L_{m-r}\le L_m$, we have
    \begin{equation*}
        \|G_{m-1}\|_{\cH}^2 \le\frac{L_m^2 \alpha}{\delta} \bbE_{\omega}[K_{m-r, r}] +  8\sigma^2. 
    \end{equation*}
Average the two inequalities over $r = 0,\dots, m-1$, 
\begin{equation*}
    A_{m-1} \le \frac{1}{m} \sum_{r=0}^{m-1} \bbE_{\omega}[\aver{\mu_r, x_r - x^{\ast}}] + \frac{\Theta_m}{\delta} \overline{K}_m,
\end{equation*}
where 
\begin{equation*}
\begin{aligned}
    \overline{K}_m &= \frac{1}{m} \sum_{r=0}^{m-1} \bbE_{\omega}[K_{m-r, r}] \\
    &\le  \frac{1}{m} \left( \frac{16}{\delta} \sum_{r=0}^{m-1} \bbE_{\omega}[\varphi(x_r; z_{r+1})] \right) + \left(34 + \frac{8\alpha}{\delta}\right) \sigma^2  + 8 \sigma^2.
\end{aligned}
\end{equation*}
By Cauchy-Schwarz and Lemma~\ref{lem: dist noise}, 
\begin{equation*}
\begin{aligned}
    \frac{1}{m} \sum_{r=0}^{m-1} \bbE_{\omega}[\aver{\mu_r, x_r - x^{\ast}}] &\le \frac{1}{m} \sum_{r=0}^{m-1} \frac{\sigma}{\sqrt{m-r}} \bbE_{\omega}[\|x_r - x^{\ast}\|_2^2]^{1/2}   \\
    &\le \frac{\sigma}{m} \sqrt{2R^2 + 26m\sigma^2/\delta} \sum_{r=0}^{m-1} (m-r)^{-1/2} \\
    &\le  \frac{2\sigma}{\sqrt{m}} \sqrt{2R^2 + 26m\sigma^2/\delta} \\
    &\le  \frac{2\sigma}{\sqrt{m}} \sqrt{2R^2} + \frac{2\sigma}{\sqrt{m}} \sqrt{ 26m\sigma^2/\delta}\\
    &\le \frac{R^2}{m} + 2\sigma^2 + \frac{2\sqrt{26}\sigma^2}{\sqrt{\delta}} \le \frac{R^2}{m} + \frac{13\sigma^2}{\delta}.
\end{aligned}
\end{equation*}
Use Lemma~\ref{lem: dist noise} again,
\begin{equation*}
\frac{1}{m} \left( \frac{16}{\delta} \sum_{r=0}^{m-1} \bbE_{\omega}[\varphi(x_r; z_{r+1})] \right) \le \frac{1}{m} \frac{16}{\delta^2} \left( 2R^2 + \frac{26 m \sigma^2}{\delta} \right) = \frac{32 R^2}{m\delta^2} + \frac{416\sigma^2}{\delta^3}.
\end{equation*}
Use the bounds $\delta\le 1$, $0<\alpha<2$ and $\Theta_m \ge 2$, we obtain
\begin{equation*}
\begin{aligned}
    A_{m-1} &\le \frac{R^2}{m} + \frac{13\sigma^2}{\delta} + \Theta_m \left( \frac{32 R^2}{m\delta^3} + \frac{474\sigma^2}{\delta^4}\right)\\
    &\le \Theta_m \left( \frac{65 R^2}{2 m\delta^3} + \frac{961 \sigma^2}{2\delta^4} \right).
\end{aligned}
\end{equation*}
The bound for $\|G_{m-1}\|_{\cH}^2$ follows similarly.
\end{proof}

\begin{theorem}
Let $0 < \alpha < 2$. Under WR for all $T\ge 1$ and under WOR for $1\le T\le n$, 
\begin{equation*}
\begin{aligned}
    \bbE_{\omega}[\widetilde{F}(x_T) - \widetilde{F}(x^{\ast})] &\le \Theta_{T+1} \left(   \frac{129 R^2}{2T\delta^3} + \frac{1909 \sigma^2}{2\delta^4}  \right), \\
    \bbE_{\omega}[\widetilde{\cG}(x_T)] &\le L_{T+1}^2 \left( \frac{136 R^2}{T\delta^3} + \frac{2002}{\delta^4}\sigma^2  \right).
\end{aligned}
\end{equation*}
\end{theorem}
\begin{proof}
    Under WR, the previous lemma with $m=T+1$ already gives the bound. Under WOR, we follow the same strategy as in Theorem~\ref{thm: no noise bd}.     
    \begin{equation}\label{eq: mean}
    \begin{aligned}
        \bbE_{\omega}[\widetilde{F}(x_T) - \widetilde{F}(x^{\ast})] &= \frac{1}{n} \left( \sum_{j=1}^{T}\bbE[ \psi(x_T; z_j) ] + \sum_{l=T+1}^n \bbE[\psi(x_T; z_l)]   \right)\\
        &= \frac{T}{n} \bbE\left[ \frac{1}{T}\sum_{j=1}^{T} \psi(x_T; z_j) \right] + \frac{n-T}{n} A_T\\
        \bbE_{\omega}[\widetilde{\cG}(x_T)] &= \frac{T}{n}\bbE\left[\frac{1}{T}\sum_{j=1}^T \|\nabla \widetilde{f}(x_T; z_j)\|^2 \right] + \frac{n-T}{n} \|G_{T}\|_{\cH}^2 .
    \end{aligned}
    \end{equation}
  We consider the same trajectory as in Theorem~\ref{thm: no noise bd}, which skips the $j$-th index. Denote the trajectory by $\{y_k^{(j)}\}_{k=0}^{T-1}$. Then
    \begin{equation*}
        y_{k}^{(j)} = x_k,\qquad 0\le k\le j-1,
    \end{equation*}
    and $x_{j} = x_{j-1} - \nabla \widetilde{f}(x_{j-1}; z_j)$. For $k\ge j+1$, 
    \begin{equation*}
        x_{k} = x_{k-1} - \nabla \widetilde{f}(x_{k-1}, z_k),\qquad y^{(j)}_{k-1} = y^{(j)}_{k-2} - \nabla \widetilde{f}(y^{(j)}_{k-2}, z_{k}).
    \end{equation*}
    Then the same argument from Theorem~\ref{thm: no noise bd} leads to 
    \begin{equation*}
    \begin{aligned}
        \|x_T - y^{(j)}_{T-1}\|_2 &\le \|\nabla \widetilde{f}(x_{j-1}, z_j)\|_2 \\
         \bbE_{\omega} [ \widetilde{f}(y_{T-1}^{(j)}; z_j) ] &= A_{T-1},\\
        \bbE_{\omega}[ \|\nabla \widetilde{f}(y_{T-1}^{(j)}; z_j)\|_2^2] &= \|G_{T-1}\|_{\cH}^2.
    \end{aligned}
    \end{equation*}
    Using $\alpha$-smoothness, 
    \begin{equation*}
    \begin{aligned}
        \widetilde{f}(x_T; z_j) &\le  \widetilde{f}(y_{T-1}^{(j)}; z_j) + \aver{  \nabla \widetilde{f}(y_{T-1}^{(j)}; z_j) , x_T - y_{T-1}^{(j)}   } + \frac{\alpha}{2} \|x_T - y_{T-1}^{(j)}\|_2^2 \\
        &\le \widetilde{f}(y_{T-1}^{(j)}; z_j) + \frac{1}{2}\| \nabla \widetilde{f}(y_{T-1}^{(j)}; z_j) \|_2^2 + \frac{1+\alpha}{2} \|x_T - y_{T-1}^{(j)}\|_2^2.
    \end{aligned}
    \end{equation*}
    Therefore, 
    \begin{equation*}
    \bbE\left[ \frac{1}{T}\sum_{j=1}^{T} \widetilde{f}(x_T; z_j) \right] \le A_{T-1} + \frac{1}{2}\|G_{T-1}\|_{\cH}^2 + \frac{1+\alpha}{2T} \sum_{j=1}^T  \|G_{j-1}\|_{\cH}^2.
    \end{equation*}
    Use Corollary~\ref{cor: grad sum} for the last summation, and apply the previous lemma for $\|G_{T-1}\|_{\cH}^2$, 
    \begin{equation*}
    \begin{aligned}
    \frac{1}{2}\|G_{T-1}\|_{\cH}^2 +  \frac{1+\alpha}{2T}  \sum_{j=1}^T  \|G_{j-1}\|_{\cH}^2 &\le \frac{1}{2}\left( \frac{32\alpha L_T^2 R^2}{T\delta^3} + \left(\frac{474\alpha L_T^2}{\delta^4} + 8\right) \sigma^2  \right)\\ &\qquad +  \frac{1+\alpha}{2T} \left(\frac{8 \alpha R^2}{\delta} + \left(\frac{104\alpha}{\delta^2} + 2\right) T\sigma^2 \right) \\
    &\le (16\alpha L_T^2 + 24)\frac{ R^2}{T\delta^3} + \left(237\alpha L_T^2 + 319 \right)\frac{\sigma^2}{\delta^4}. 
    \end{aligned}
    \end{equation*}
    And both $A_{T-1}, A_T \le \Theta_{T+1} \left( \frac{65 R^2}{2 T\delta^3} + \frac{961 \sigma^2}{2\delta^4} \right)$. Since $\Theta_m \ge 6 + \frac{\alpha}{2} L_m^2$ for $m\ge 2$, 
    \begin{equation*}
    \bbE_{\omega}\left[ \frac{1}{T}\sum_{j=1}^{T} \widetilde{f}(x_T; z_j) \right] \le \Theta_{T+1} \left(   \frac{129 R^2}{2T\delta^3} + \frac{1909 \sigma^2}{2\delta^4}  \right).   
    \end{equation*}
Thus, by~\eqref{eq: mean},
\begin{equation*}
\bbE_{\omega}[\widetilde{F}(x_T) - \widetilde{F}(x^{\ast})] \le \Theta_{T+1} \left(   \frac{129 R^2}{2T\delta^3} + \frac{1909 \sigma^2}{2\delta^4}  \right).   
\end{equation*}
Similarly, for the gradient bound, we use $\alpha$-smoothness,  
\begin{equation*}
\begin{aligned}
  \|\nabla \widetilde{f}(x_T; z_j) \|_2^2 &\le  2\|\nabla \widetilde{f}(y_{T-1}^{(j)}; z_j)\|_2^2 +  2\|\nabla \widetilde{f}(x_T; z_j) - \nabla \widetilde{f}(y_{T-1}^{(j)}; z_j)\|^2_2 \\ & \le 2\|\nabla \widetilde{f}(y_{T-1}^{(j)}; z_j)\|_2^2 + 2\alpha^2 \|x_T - y_{T-1}^{(j)} \|_2^2.
\end{aligned}
\end{equation*}
Sum over $j=1,\dots, T$, we use $0< \alpha < 2$,  $L_T \ge 4$, the bound from Lemma~\ref{lem: n bd}, and Corollary~\ref{cor: grad sum},
\begin{equation*}
\begin{aligned}
    \bbE_{\omega}\left[\frac{1}{T}\sum_{j=1}^T \|\nabla f(x_T; z_j)\|^2 \right] &\le 2 \|G_{T-1}\|_{\cH}^2 + \frac{2\alpha^2 }{T} \sum_{j=1}^T  \|G_{j-1}\|_{\cH}^2 \\
    &\le \left(64\alpha L_T^2 + 16\alpha^3\right) \frac{R^2}{T\delta^3}+  \frac{2\alpha^2}{T} \left( \frac{8 \alpha R^2}{\delta} + \left(\frac{104\alpha}{\delta^2} + 2\right) T\sigma^2  \right)\\
    &\le L_T^2 \left( \frac{136 R^2}{T\delta^3} + \frac{2002}{\delta^4}\sigma^2  \right).
\end{aligned}
\end{equation*}
\end{proof}
Removing the normalization $F = \eta^{-1} \widetilde{F}$, and replace with $\sigma_{\ast}^2 = \sigma^2 \eta^{-2} $, we obtain 
\begin{equation*}
\bbE_{\omega}[{F}(x_T) - {F}(x^{\ast})] \le \Theta_{T+1} \left(   \frac{129 R^2}{2T\eta \delta^3} + \frac{1909 \eta \sigma_{\ast}^2}{2 \delta^4}  \right).
\end{equation*}
Since the factor $\Theta_{T+1} = \cO(\ell_T^3/\lambda_T)$, we have proved Theorem~\ref{thm: main}. 
\section{Conclusion}\label{sec: conc}
We proved a last-iterate bound for SGD with and without replacement in both interpolation and low-noise regimes. Our bound slightly improves the earlier result $(\log T)^3/T$ in~\citet{} by a factor of $\log\log T$. Further improvement would require additional control from the trajectories, potentially through a quantitative interaction of the gradient alignment from the dyadic argument and the contraction along shared trajectories.

\section*{Acknowledgments}
YZ was partially funded by NSF grant DMS-2309530.

\bibliographystyle{plainnat}
\bibliography{main4}

\end{document}